\documentclass{article} 
\usepackage{iclr2027_conference,times}

\usepackage{amsmath,amsfonts,bm}

\def\eqref#1{equation~\ref{#1}}

\def\1{\bm{1}}

\DeclareMathAlphabet{\mathsfit}{\encodingdefault}{\sfdefault}{m}{sl}
\SetMathAlphabet{\mathsfit}{bold}{\encodingdefault}{\sfdefault}{bx}{n}

\usepackage{hyperref}
\usepackage{xcolor}
\usepackage{colortbl}
\usepackage{subcaption}
\usepackage{pdflscape}
\usepackage{longtable}
\usepackage{url}
\usepackage{arydshln}
\usepackage{algorithm}
\usepackage{makecell}
\usepackage{needspace}
\usepackage{wrapfig}
\usepackage{algpseudocode}
\hypersetup{hidelinks}
\definecolor{linkblue}{RGB}{0,70,140}

\hypersetup{
    colorlinks=true,
    linkcolor=linkblue,
    citecolor=linkblue,
    urlcolor=linkblue
}
\usepackage{algpseudocode}
\usepackage{caption}
\newtheorem{theorem}{Theorem}
\newtheorem{lemma}{Lemma}
\usepackage{graphicx}
\usepackage{booktabs}

\usepackage[most]{tcolorbox}
\definecolor{exnavy}{HTML}{0F3D5E}
\definecolor{exback}{RGB}{247,250,251}
\newtcolorbox{promptbox}[1][]{
    enhanced,
    colback=gray!8,
    colframe=gray!50,
    boxrule=0.5pt,
    arc=2pt,
    left=6pt,
    right=6pt,
    top=8pt,
    bottom=6pt,
    title=#1,
    fonttitle=\bfseries\small,
    coltitle=white,
    attach boxed title to top left={
        xshift=8pt,
        yshift=-4pt
    },
    boxed title style={
        colback=exnavy,
        colframe=exnavy,
        boxrule=0.5pt,
        arc=2pt,
        left=4pt,
        right=4pt,
        top=1pt,
        bottom=1pt
    }
}

\usepackage[most]{tcolorbox}
\defcitealias{workshop2022bloom}{BigScience et al.}

\usepackage{xcolor}

\definecolor{PromptBG}{RGB}{247,248,250}
\definecolor{PromptAccent}{RGB}{155,163,175}
\definecolor{PromptTitle}{RGB}{82,88,99}

\newtcolorbox{promptdisplay}{
    enhanced,
    colback=PromptBG,
    frame hidden,
    borderline west={1.8pt}{20pt}{PromptAccent},
    arc=2pt,
    left=30pt,
    right=30pt,
    top=4pt,
    bottom=4pt,
    before skip=5pt,
    after skip=5pt,
}

\newtcolorbox{examplebox}[1][]{enhanced, breakable,
  colback=exback, colframe=exnavy!30, boxrule=0.6pt, arc=2pt,
  left=7pt, right=7pt, top=8pt, bottom=6pt,
  fonttitle=\bfseries\small, coltitle=white, title=#1,
  attach boxed title to top left={xshift=8pt, yshift=-4pt},
  boxed title style={colback=exnavy, colframe=exnavy, boxrule=0pt, arc=2pt,
                     left=5pt, right=5pt, top=1.5pt, bottom=1.5pt}}
\newcommand{\exrow}[2]{\par\noindent{\footnotesize\textsc{#1}}\; #2\par}

\providecolor{LightGray}{HTML}{F5F7F9}
\providecolor{SectionGray}{HTML}{E9EDF2}

\title{Nash Equilibrium Text: A Game-Theoretic \\ Decoding Framework for Text Generation}

\author{
Alireza Jafari \\
Department of Computer Science \\
University of Virginia \\
Charlottesville, VA, USA \\
\texttt{jrp5td@virginia.edu} \\
\And
Arman Adibi \\
School of Computer and Cyber Sciences \\
Augusta University \\
Augusta, GA, USA \\
\texttt{aadibi@augusta.edu} \\
\AND
Mohammad Ghavamzadeh \\
Qualcomm AI Research \\
San Diego, CA, USA \\
\texttt{mohammad.ghavamzadeh51@gmail.com} \\
\And
Hadi Daneshmand \\
Department of Computer Science \\
University of Virginia \\
Charlottesville, VA, USA \\
\texttt{dhadi@virginia.edu}
}

\iclrfinalcopy 
\usepackage{enumitem}
\begin{document}
\newcommand{\ravg}{\mathrm{r_{avg}}}

\maketitle
\lhead{}
\renewcommand{\headrulewidth}{0pt}

\begin{abstract}
Text revision has become an integral component of large language models. This paper formulates revision such that it admits a Nash equilibrium: Token positions are players, vocabulary items are actions, and each player's utility is the language model's log conditional probability. We motivate the revision by showing that Nash equilibria can have exponentially higher likelihood than autoregressive outputs as the sequence length grows. We further propose \emph{Nash decoding}, an algorithm that reaches an $\varepsilon$-Nash equilibrium in $O(1/\varepsilon)$ time given access to the joint probability of tokens conditioned on a prompt. In practice, we run Nash decoding using conditional probability estimates from large language models and evaluate the resulting equilibria on question-answering benchmarks. On CLAPNQ, PubMedQA, and CoQA, Nash equilibria obtained from masked language models achieve higher F1 and ROUGE scores than autoregressive models up to $18\times$ larger, without any fine-tuning or retraining, at the cost of additional test-time computation. Code is available at \url{https://github.com/alireza-jafari/Nash-Decoding}.
\end{abstract}

\section{Introduction}
\label{sec:introduction}
“The first draft of anything is shit,” Ernest Hemingway reportedly remarked
\citep{samuelson1984with}. The first draft of this manuscript was no exception.
Accomplished writers, at least before the
advent of LLMs, have devoted considerable effort to revision. Hemingway,
for instance, rewrote the final page of \emph{A Farewell to Arms} 39 times~\citep{plimpton1958hemingway}.
This raises a scientific question: 
\begin{center}
\emph{When does the revision terminate, and what is the
equilibrium text that editing converges to?}
\end{center}

Revision has become the core in text generation with language models where the first draft often is the standard de facto autoregressive text obeying
\begin{align} \label{eq:AR}
x_{i+1}^{(\text{AR})}= \arg\max_{x \in \mathcal{V}} \; p(x \mid x_{\leq i}^{(\text{AR})}; \text{prompt}), 
\end{align}
where $p$ denotes the probability function  \citep{radford2019language} and $x_i$ are tokens selected from a vocabulary $\mathcal{V}$. Autoregressive generation calls for revision because each token is produced
without access to the tokens that follow. Existing approaches to revision fall
into four broad families: (1) \emph{remasking} tokens in diffusion and masked
language models~\citep{ghazvininejad2019maskpredict, sahoo2024mdlm, shi2024simplified};
(2) \emph{reinforcement learning} with external
feedback~\citep{ouyang2022training, openai2024o1, guo2025deepseek};
(3) \emph{Markov chain Monte Carlo (MCMC)}
sampling~\citep{karan2026reasoning, zhao2026learning, xu2025energy, wang2019bert}; and
(4) \emph{hybrid} decoding
techniques~\citep{xia2017deliberation, lee2018iterative, gu2019levenshtein, welleck2023selfcorrect}.
We focus on revision driven solely by conditional probabilities without external
feedback. Such revision need not reach an equilibrium; the text may continue to
change indefinitely. We therefore formulate revision as a game that admits an
equilibrium.

\begin{figure}[t]
    \centering
    \includegraphics[width=1\linewidth]{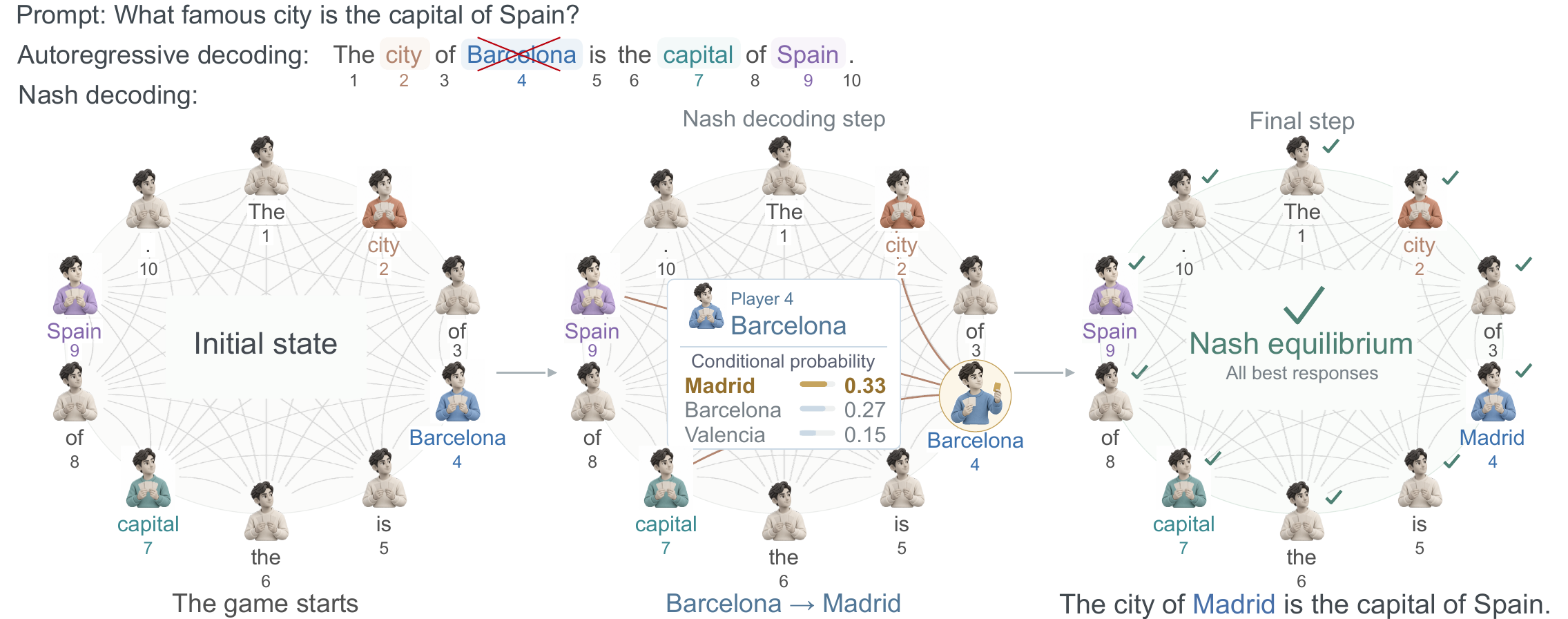}
   \caption{\small \textbf{Revision Game.} Each token position is a player whose choice is evaluated given all context tokens. Autoregressive decoding leaves earlier tokens fixed, whereas the tokens are
revised in the revision game. We used GPT-2 small for conditional probability estimates. At a Nash equilibrium, no player can increase its conditional probability by changing its token. Details are reported in Appendix~\ref{app:spain-token-game}.}
    \label{fig:token-game}
\end{figure}

\emph{Revision game}: Given a prompt, $n$ players
are the positions $i\in[n]$, and each player's action is a token
from $\mathcal V$ and player $i$ receives utility
$\log p(x_i \mid x_{-i}; \mathrm{prompt})$, where $x_{-i}$ denotes all tokens except $x_i$.
Since $p(x_i \mid x_{-i}; \mathrm{prompt}) = p(x_1, \dots, x_n \mid \mathrm{prompt}) / p(x_{-i} \mid \mathrm{prompt})$
and the denominator does not depend on $x_i$, this is an exact potential game with potential
${\log p(x_1, \dots, x_n \mid \mathrm{prompt})}$~\citep{monderer1996potential}. Being finite, the
game admits at least one pure Nash equilibrium
$x^{(\mathrm{nash})} = (x_1^{(\mathrm{nash})}, \ldots, x_n^{(\mathrm{nash})})$, which satisfies
\begin{align} \label{eq:nash_equ}
\forall i \in [n]: \quad x_i^{(\mathrm{nash})} = \arg\max_{x \in \mathcal{V}} \; p\big(x \mid x_{-i}^{(\mathrm{nash})}; \mathrm{prompt}\big).
\end{align}
In other words, a sequence is a \emph{Nash equilibrium} when no single position can
increase its conditional probability by changing its token. Comparing to Equation~\ref{eq:AR}, the above conditioning
 is on $x_{-i}$ rather than $x_{<i}$.

As an example, we simulate this revision game between tokens using  GPT-2 prompted by the
question ``What famous city is the capital of Spain?''. Figure~\ref{fig:token-game} shows revisions of the first draft generated by autoregressive decoding that is ``The city of Barcelona is the capital of Spain.'' As the game proceeds, the token ``Barcelona'' is revised to ``Madrid'',
increasing the likelihood of the full generated text by a factor of $1.26$\footnote{In this example, the increase in probability happens to coincide with
improved factual correctness; in general, higher probability does not ensure correctness.}. We will show that Nash equilibria
can attain substantially higher likelihood than autoregressive generations:
the ratio $\frac{p(x^{(\mathrm{nash})} \mid \mathrm{prompt})}{
p(x^{(\mathrm{AR})} \mid \mathrm{prompt})}$ can grow exponentially in $n$.

Computing an exact pure Nash equilibrium in general potential games is PLS-complete \citep{fabrikant2004complexity}.
In contrast, we propose a revision mechanism called Nash decoding that obtains an $\epsilon$-Nash equilibrium in
$O\!\left( 1 / \varepsilon \right)$ time.  Furthermore, the conditional probability of revised text monotonically increases through the revision.

Empirically, we show that masked language models can be used to compute a Nash equilibrium. Thus, we evaluate the question-answering performance of Nash equilibria against autoregressive generation on question-answering benchmarks. At the cost of additional computation, iterative Nash updates allow a masked language model to surpass the F1 and ROUGE of autoregressive models with 18 times as many parameters on CoQA \citep{reddy-etal-2019-coqa}, PubMedQA~\citep{jin-etal-2019-pubmedqa}, and CLAPNQ \citep{rosenthal-etal-2025-clapnq} datasets. We also report negative results showing that scores of equilibrium vary and depend on the initialization and the underlying model.

\section{Text Generation as a Game}
\label{sec:game}

Why a Nash equilibrium? We motivate finding a Nash equilibrium by proving that its likelihood can be significantly larger than that of autoregressive text.

\begin{lemma}[Separation between greedy AR decoding and Nash equilibrium]
\label{lem:ar-nash-separation}
For every positive even integer $n$, there exists a strictly positive autoregressive
distribution on $\{0,1\}^n$ such that
\[
\frac{p(x^{(\mathrm{nash})})}{p(x^{(\mathrm{AR})})}
=
\exp(\gamma n),
\qquad
\gamma:=\frac{1}{2}\log\!\left(\frac{539}{289}\right)
\approx 0.31164.
\]

\end{lemma}
The proof is postponed to Appendix~\ref{app:proof-ar-nash-separation}.
While the above result is proven from a specific joint probability function, we will later show that for every probability function there exists at least one Nash equilibrium with higher likelihood than the autoregressive output (if it is not an equilibrium).

\subsection{Approximate Equilibrium Is Attainable}
\label{sec:equilibrium-convergence}
We study when \emph{revision terminates}. We prove that there exists a revision
process that terminates at an approximate Nash equilibrium. To make this notion
precise, we first define the \emph{token gap} at position $i$ as the largest
increase in conditional probability that a single token replacement at that
position can achieve: \begin{equation}
g_i(x)
=
\max_{v\in\mathcal V}
\;p(v\mid x_{-i};\text{prompt})
-
p(x_i\mid x_{-i};\text{prompt}).
\label{eq:token-gap}
\end{equation}
The \emph{Nash gap} is the largest improvement available at any position:
$G(x)=\max_{i\in[n]}g_i(x)$.
The equilibrium condition for $x^{(nash)}$ in Equation~\ref{eq:nash_equ} is therefore equivalent to $G(x^{(nash)})=0$: no position can improve its utility through a unilateral token replacement. More generally, for a tolerance $\varepsilon\geq0$, $x$ is an \emph{additive $\varepsilon$-Nash equilibrium} if $G(x)\leq\varepsilon$: no single-token replacement improves its position's conditional probability by more than $\varepsilon$.

Recall the formulation of conditional density as
\begin{equation}
p(x_i\mid x_{-i};\text{prompt})
=
\frac{
p(x\mid\text{prompt})
}{
p(x_{-i}\mid\text{prompt})
}.
\label{eq:full-context-conditional}
\end{equation}
The denominator is positive and unchanged by a unilateral deviation. Thus, whenever a player increases its utility, the joint probability also increases. In game-theoretic terms, the joint probability is an \emph{ordinal potential} \citep{monderer1996potential}. This property yields a bound on the number
of improving token refinements.

We define \emph{Nash decoding} as the maximum-gap best-response procedure in Algorithm~\ref{alg:nash-decoding}.
Given a tolerance $0<\varepsilon\leq1$, start from any completed
sequence $x^{(0)}\in\mathcal V^n$. At iteration $k\geq0$, compute
$g_i(x^{(k)})$ for every position $i\in[n]$. If
$G(x^{(k)})\leq\varepsilon$, return $x^{(k)}$. Otherwise, select a position
with the largest gap and replace its token with the best response,
leaving all other tokens unchanged.
It terminates when the maximum Nash gap is at most $\varepsilon$, yielding an $\varepsilon$-Nash equilibrium.

\enlargethispage{\baselineskip}

\begin{algorithm}[H]
\caption{Nash Decoding}
\label{alg:nash-decoding}
\begin{algorithmic}[1]
\Require Prompt, initial sequence $x^{(0)}\in\mathcal V^n$, tolerance $0<\varepsilon\leq 1$
\Ensure Sequence $x$ satisfying $G(x)\leq\varepsilon$
\State $k\leftarrow 0$
\While{$G(x^{(k)})>\varepsilon$}
\State Compute $g_i(x^{(k)})$ for all $i\in[n]$
\State $G(x^{(k)})\leftarrow\max_{i\in[n]} g_i(x^{(k)})$
\State Select
\(    i_k\in\arg\max_{i\in[n]} g_i(x^{(k)})
   \)
\State Set
\(    x_{i_k}^{(k+1)}
    \in
    \arg\max_{v\in\mathcal V}
    p\!\left(x_{i_k}=v
    \mid x_{-i_k}^{(k)};\mathrm{prompt}\right)
   \)
\State Set $x_i^{(k+1)}\leftarrow x_i^{(k)}$ for all $i\neq i_k$
\State $k\leftarrow k+1$
\State \textbf{Break if} $x^{(k)}$ is visited before
\EndWhile \\
\Return $x^{(k)}$
\end{algorithmic}
\end{algorithm}
\vspace{-3mm}

The next theorem shows that Nash decoding monotonically increases the sequence probability in each iteration and reaches an $\varepsilon$-Nash equilibrium in a polynomial number of iterations.

\begin{theorem}[Convergence Rate]
\label{thm:convergence-rate}
Let $0<\varepsilon\leq1$ and let $p_0=p(x^{(0)}\mid\text{prompt})>0$ be the initial sequence probability. Every performed update of the above dynamics satisfies
\begin{equation}
\frac{p(x^{(k+1)}\mid\text{prompt})}
{p(x^{(k)}\mid\text{prompt})}
\geq 1+\underbrace{G(x^{(k)})}_{>0}.
\label{eq:monotone-improvement}
\end{equation}
The dynamics terminate at an additive $\varepsilon$-Nash equilibrium after $K_\varepsilon$ token replacements, with
\begin{equation}
K_\varepsilon\leq
\frac{\log(1/p_0)}{\log(1+\varepsilon)}
\leq\frac{2\log(1/p_0)}{\varepsilon}.
\label{eq:convergence-rate}
\end{equation}
\end{theorem}

\paragraph{Proof.}
Let $p_k=p(x^{(k)}\mid\text{prompt})$. Equation~\ref{eq:full-context-conditional} and maximum-gap selection imply
\[
p_{k+1}-p_k
=p(x_{-i_k}^{(k)}\mid\text{prompt})G(x^{(k)})
\geq p_kG(x^{(k)}),
\]
since the marginal probability is at least the probability of the completed sequence. This proves Equation~\ref{eq:monotone-improvement}. Each performed update has $G(x^{(k)})>\varepsilon$, so any $K$ performed updates satisfy
\[
1\geq p_K\geq p_0(1+\varepsilon)^K.
\]
Taking logarithms bounds the number of updates; hence the dynamics must terminate with $G(x)\leq\varepsilon$. Finally, $\log(1+\varepsilon)\geq\varepsilon/2$ for $0<\varepsilon\leq1$. \hfill$\square$

The bound depends logarithmically on the inverse initial probability: a more probable starting sequence yields a tighter guarantee. To distinguish replacements from their computational cost, count one \emph{conditional query} as obtaining $p(\cdot\mid x_{-i};\text{prompt})$ over the entire vocabulary at one position. Recomputing all gaps after each replacement requires
$
O\!\left(n\log(1/p_0) \varepsilon^{-1} \right)
\label{eq:conditional-query-complexity}
$ density evaluations, including the final scan that certifies termination.

A consequence of this result is: for every $x^{\text{(AR)}}$ in Equation~\ref{eq:AR}, there exists a Nash equilibrium whose likelihood is higher as long as it is not an equilibrium. This follows because Nash decoding, initialized at the autoregressive output, monotonically increases the likelihood until it reaches an equilibrium.

\subsection{Nash Decoding with Pretrained Language Models}
\label{sec:nash-decoding}

Nash decoding relies on the conditional probability $p(v \mid x_{-i}; \text{prompt})$ of token $v$ at position $i$ given the rest of the sequence. This quantity can be estimated with either autoregressive or masked language models.

\paragraph{Masked language models.}
A masked language model (MLM), such as BERT, provides the conditional
estimates needed for Nash decoding. For each position $i$, we replace
$x_i$ with \texttt{[MASK]} while retaining the prompt and all other
tokens. The model produces a logit $z_{\theta,i}(v)$ for each candidate
token $v\in\mathcal V$ at the masked position. Applying softmax gives
\[
q_\theta(v\mid x_{-i};\text{prompt})
=
\frac{\exp(z_{\theta,i}(v))}
{\sum_{w\in\mathcal V}\exp(z_{\theta,i}(w))},
\]
where $w$ ranges over the vocabulary and the logits depend on the
masked input. Because the model is pretrained to reconstruct masked
tokens from their surrounding context, we use this distribution as
an estimate of $p(v\mid x_{-i};\text{prompt})$
\citep{devlin2019bert}. 

\paragraph{Autoregressive  models.}
Nash decoding can also use an AR model, as illustrated by the motivating example in Section~\ref{sec:introduction}. An AR model defines a joint distribution $p_\theta^{\mathrm{AR}}(x\mid\text{prompt})$ as the product of its left-to-right token probabilities, so its full-context conditional is
\begin{equation}
q_\theta^{\mathrm{AR}}(v\mid x_{-i};\text{prompt})
=
\frac{
p_\theta^{\mathrm{AR}}(x_{<i},v,x_{>i}\mid\text{prompt})
}{
\displaystyle\sum_{w\in\mathcal V}
p_\theta^{\mathrm{AR}}(x_{<i},w,x_{>i}\mid\text{prompt})
}.
\label{eq:ar-full-conditional}
\end{equation}
But computing the above conditional probability is expensive because each replacement can change subsequent token probabilities, requiring up to $|\mathcal V|$ candidate suffix evaluations per position.

\paragraph{The challenge of joint density estimation.} The conditional distributions supplied by a masked language
model need not be compatible with a common joint distribution
\citep{torroba-hennigen-kim-2023-deriving} with a joint probability $p$ where
\begin{equation}
\begin{aligned}
p(x \mid \text{prompt})
&= p(x_i \mid x_{-i}; \text{prompt})\, p(x_{-i} \mid \text{prompt}) \\
&= p(x_j \mid x_{-j}; \text{prompt})\, p(x_{-j} \mid \text{prompt})
\end{aligned}
\label{eq:joint_property}
\end{equation}
holds for every sequence $x$ and every pair of positions $i$ and $j$. 
Consequently, the convergence guarantee
in Theorem~\ref{thm:convergence-rate} does not necessarily hold for the conditional probability estimates; hence, we empirically validate the convergence to Nash equilibrium.

\Needspace{10\baselineskip}

\subsection{Empirical Validation of  Theorem~\ref{thm:convergence-rate}}
\label{sec:empirical-dynamics}

\begin{wrapfigure}{r}{0.37\textwidth}
    \vspace{-30pt}
    \centering

    \begin{subfigure}{\linewidth}
        \centering
        \includegraphics[width=0.93\linewidth]{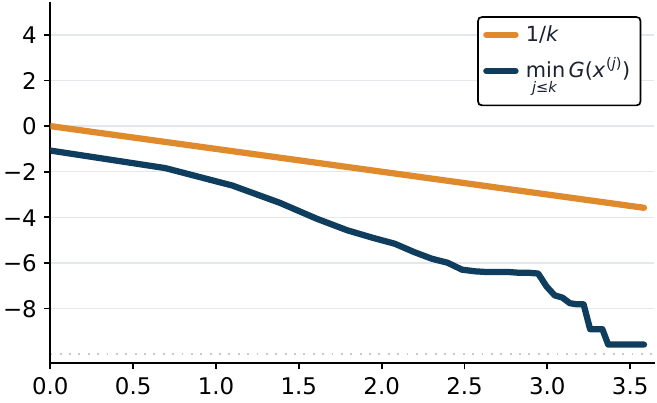}
        \caption{\small Convergence Test}
        \label{fig:nash-gap-bound}
    \end{subfigure}

    \vspace{4pt}

    \begin{subfigure}{\linewidth}
        \centering
        \includegraphics[width=0.96\linewidth]{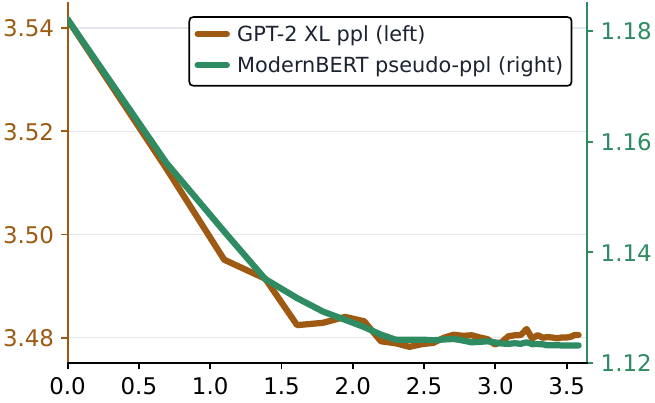}
        \caption{\small Monotonicity Test}
        \label{fig:ppl-refinement}
    \end{subfigure}
    \vspace{-15pt}
\caption{\small
        Convergence of Nash decoding on WikiText-103.
        (a) Running minimum of the Nash gap,
        compared with a $1/k$ reference curve. The reference
        indicates a rate of decay, aligned with Theorem~\ref{thm:convergence-rate}. Both axes are log-transformed.
        (b) GPT-2 XL continuation perplexity 
        and ModernBERT pseudo-perplexity. The x-axis is log-transformed.
    }
    \vspace{-25pt}
    \label{fig:theory-validation}
\end{wrapfigure}

We experimentally validate (1) the convergence rate and (2) likelihood monotonicity in Theorem~\ref{thm:convergence-rate}. Using ModernBERT
\citep{warner2025modernbert}, we generate 64-token continuations for 500
prompts from WikiText-103 \citep{merity2017pointer}. Each continuation starts from 64 \texttt{[MASK]} tokens and repeatedly replaces the maximum-gap token with its best response (see details
in Appendix~\ref{app:wikitext}).

\emph{(1) Convergence test:} Figure~\ref{fig:nash-gap-bound} compares the empirical convergence rate of the Nash gap with the linear rate established in Theorem~\ref{thm:convergence-rate}. We track the rate only when all tokens are unmasked (see Appendix~\ref{app:equilibrium_diagnostic}). The Nash Gap remains below the theoretical rate $1/k$. The plateaus near the end are due to the fact that most texts have already reached equilibrium at a faster rate compared to the theoretical rate $O(1/k)$. Notably, this observation holds only when all tokens are unmasked.

\emph{(2) Monotonicity test:} To verify the likelihood increase, we report the perplexity of text obtained across iterations of Nash decoding. Figure~\ref{fig:ppl-refinement} shows ModernBERT pseudo-perplexity and perplexity under an external autoregressive model, GPT-2 XL~\citep{radford2019language}. Both scores decrease with iterations, indicating an increase in likelihood. Although the likelihood increases does not necessarily hold for conditional probability estimates, as discussed in the last section, the observed decrease in pseudo-perplexity is consistent with the likelihood increase in Theorem~\ref{thm:convergence-rate}.

\section{Related Work}
\label{app:related}

Nash decoding, introduced in this work, links to a family of decoding methods which we categorize into three main groups: (1) multi-pass decoders, (2) parallel decoders, and (3) speculative decoders.

\emph{(1) Multi-pass decoders.}
This work is inspired by a rich body of prior work on refining autoregressive text generation to improve the likelihood of the generated text. \citet{xia2017deliberation} address this limitation with a first-pass autoregressive decoder that produces a complete draft and a trained second-pass decoder that refines it. The first generator might itself prefer to refine any token once the draft is complete. Self-Refine instead uses the same model to generate and revise its output \citep{madaan2023selfrefine}, but its revisions target task-specific quality criteria rather than certifying token-level equilibrium. \cite{zhou2026dllm} define a competitive MCMC \emph{sampling} (remasking) mechanism that bridges the gap between masked and autoregressive language models.

\emph{(2) Parallel decoders.}
Like masked diffusion language models \citep{sahoo2024mdlm,shi2024simplified}, Nash decoding iteratively revises tokens. However, diffusion language models focus on parallel text generation \citep{austin2021structured,nie2025llada}, whereas our procedure for finding a Nash equilibrium is sequential, updating one token at a time. This reflects a difference in objectives: rather than computational efficiency of decoding, we focus on characterizing the likelihood properties of the Nash equilibrium.

\emph{(3) Speculative decoders.} Speculative decoding accelerates autoregressive generation by using a draft model to propose tokens and a target model to verify them~\citep{leviathan2023speculative,chen2023speculative}. In SpecDec, a drafted token is revised if it falls outside the verifier’s top tokens or if its log-likelihood is more than a threshold below that of the verifier’s top prediction. At the first such position, the verifier replaces the token with its own prediction and discards the remaining draft \citep{xia2023speculative}.
Nash decoding differs from these methods in two ways: (1) it selects tokens for revision based on the Nash gap, and (2) it provably converges to an approximate Nash equilibrium.

A growing body of work studies game-theoretic interactions among language-model agents and response policies~\citep{zhang2025strategic,jacob2024consensus,munos2024nash,hagen2026wasserstein}, including test-time methods for human-value alignment~\citep{zhang2024truth,tamine2025data}. Nash learning from human feedback~\citep{munos2024nash}, for example, seeks an equilibrium between competing response policies. In contrast, we formulate decoding itself as a game in which positions within a single generated sequence are players and tokens are their actions.

\section{Empirical Results and Benchmarking}
\label{sec:experiments}

We assess the question-answering performance of Nash decoding compared to existing autoregressive and
masked diffusion decoding. 
For conditional probability estimates, we used four pretrained models:
ModernBERT-Large (395M)~\citep{warner2025modernbert},
RoBERTa-Large (355M)~\citep{liu2019roberta},
mmBERT-base~\citep{marone2025mmbert}, and 
Ettin-400M (396M)~\citep{weller2025ettin}.

\paragraph{Datasets.}
We used three standard question-answering benchmarks akin to prior works~\citep{reddy-etal-2019-coqa,dong2019unified,rosenthal-etal-2025-clapnq,luo2022biogpt,huang-etal-2023-diffusion}: 
\begin{itemize}[leftmargin=1.5em, nolistsep]
    \item[(1)] CLAPNQ~\citep{rosenthal-etal-2025-clapnq}
pairs a Natural Questions passage with either a cohesive long-form answer or no answer at
all; we use the $300$ \emph{answerable} questions from its development split (mean answer length $63.8$ tokens). 
\item[(2)] From PubMedQA~\citep{jin-etal-2019-pubmedqa}, we use the $500$ PubMedQA-Labeled test questions, where the model must infer the abstract's withheld conclusion rather than extract or repeat text from the input (mean answer length $48.3$ tokens).
\item[(3)] From CoQA~\citep{reddy-etal-2019-coqa}, we use the $1{,}814$ development
questions whose reference answer contains at least 5 tokens (using ModernBERT tokenizer); this threshold
excludes bare \emph{yes}/\emph{no} and other very short answers, leaving enough positions for meaningful token interactions (mean answer length $7.5$ tokens) \footnote{To verify that the $\geq 5$-token filter does not drive the CoQA results, we also evaluate on the full development set, including all questions; see Appendix~\ref{app:qa-results}.}.
\end{itemize}
Appendix~\ref{app:datasets} describes how the long-answer questions were selected and reports their statistics.

\paragraph{Protocol.}
We use a fixed prompt format for all models within each dataset. For CLAPNQ and PubMedQA, the passage and question are followed by the soft instruction ``This question is answered completely with evidence from the passage. Answer:'' and an \emph{Answer:} prefix. For CoQA, which consists of conversational question--answer turns, we instead use the standard \emph{Q:} and \emph{A:} dialogue format following the original evaluation setup~\citep{radford2019language}. Across all datasets, every model receives the same oracle answer-length budget $T$, computed from the reference answer using that model's tokenizer. End-of-sequence and structural tokens are banned, so every model emits exactly $T$ tokens. For Nash decoding, the masked model is used without additional training: we append a canvas of $T$ \texttt{[MASK]} tokens to the prompt and iteratively update tokens using the model's full-context conditionals until the Nash gap satisfies $G(x)\leq 10^{-4}$. We report SQuAD-style token F1~\citep{rajpurkar-etal-2016-squad}, scored against the most favorable reference, together with ROUGE-L and ROUGE-Lsum~\citep{lin-2004-rouge}. Appendix~\ref{app:config} provides the complete protocol.

\subsection{Question-Answering Performance}

Table~\ref{tab:main_results} reports the question-answering performance of masked
language models with and without Nash decoding, alongside strong autoregressive
baselines. With Nash decoding, masked language models become effective zero-shot
question answerers, outperforming autoregressive models $18\times$ their size in
F1, ROUGE-L, and ROUGE-Lsum. These gains come at the cost of additional test-time
computation, which we analyze in Section~\ref{sec:discussion}.

\begin{table*}[h]
\centering
\setlength{\tabcolsep}{2.8pt}
\small
\caption{\small 
Results across three question-answering benchmarks under a shared
evaluation protocol.
\textbf{Size} denotes the number of model parameters; for Nash decoding, it denotes the masked
model size.
Best results are \textbf{bolded} and second-best results are \underline{underlined}. The complete benchmark results are reported in Appendix~\ref{app:qa-results}.}

\label{tab:main_results}

\begin{tabular}{llccc ccc ccc}
\toprule
& &
\multicolumn{3}{c}{\textbf{CoQA}} &
\multicolumn{3}{c}{\textbf{PubMedQA}} &
\multicolumn{3}{c}{\textbf{CLAPNQ}} \\
\cmidrule(lr){3-5}
\cmidrule(lr){6-8}
\cmidrule(lr){9-11}
\textbf{Model} & \textbf{Size}
& F1 & R-L & R-Ls
& F1 & R-L & R-Ls
& F1 & R-L & R-Ls \\
\midrule

\rowcolor{SectionGray}
\multicolumn{11}{l}{\textbf{One-shot masked LMs (not fine-tuned)}} \\
\rowcolor{white}
RoBERTa-L~\citep{liu2019roberta}
& 355M
& 28.00 & 31.45 & 31.46
& 7.99 & 8.21 & 8.74
& 5.94 & 10.38 & 10.59 \\

\rowcolor{LightGray}
ModernBERT-L~\citep{warner2025modernbert}
& 395M
& 39.54 & 42.92 & 42.93
& 5.93 & 7.95 & 8.34
& 7.03 & 12.05 & 12.66 \\

Ettin-400m~\citep{weller2025ettin}
& 396M
& 43.87 & 46.22 & 46.24
& 4.03 & 4.55 & 4.76
& 6.97 & 11.78 & 12.41 \\




\midrule
\rowcolor{SectionGray}
\multicolumn{11}{l}{\textbf{Autoregressive models}} \\

\rowcolor{white}
OPT-350M~\citep{zhang2022opt}
& 331M
& 27.16 & 29.86 & 29.94
& 23.43 & 20.43 & 21.41
& 33.21 & 32.14 & 33.39 \\

\rowcolor{LightGray}
SmolLM2-360M~\citep{allal2025smollm2}
& 362M
& 52.77 & 54.30 & 54.32
& 21.56 & 19.71 & 20.57
& 32.62 & 31.02 & 32.66 \\

GPT-2 L~\citep{radford2019language}
& 774M
& 42.61 & 44.31 & 44.38
& 22.09 & 19.85 & 20.84
& 33.06 & 31.50 & 33.68 \\

\rowcolor{LightGray}
GPT-2 XL~\citep{radford2019language}
& 1.5B
& 46.83 & 48.26 & 48.35
& 23.78 & \underline{20.77} & \underline{22.00}
& 30.58 & 29.51 & 30.73 \\

BLOOM~(\citetalias{workshop2022bloom},~\citeyear{workshop2022bloom}) 
& 3B
& 56.69 & 56.25 & 56.62
& 22.60 & 19.53 & 21.46
& 34.00 & \underline{32.18} & \underline{34.31} \\

\rowcolor{LightGray}
OLMo-1.7~\citep{groeneveld2024olmo}
& 7B
& 51.44 & 53.53 & 53.57
& 19.67 & 17.71 & 20.09
& 30.12 & 28.63 & 32.15 \\

\midrule
\rowcolor{SectionGray}
\multicolumn{11}{l}{\textbf{Nash decoding with same masked LMs}} \\
\rowcolor{white}
RoBERTa-L
& 355M
& 41.35 & 44.02 & 44.04
& \textbf{24.99} & \textbf{21.17} & \textbf{22.69}
& 25.25 & 25.59 & 25.17 \\

\rowcolor{LightGray}
ModernBERT-L
& 395M
& \underline{58.12} & \underline{58.71} & \underline{58.71}
& \underline{24.28} & 20.33 & 21.36
& \textbf{38.04} & \textbf{34.79} & \textbf{37.42} \\

Ettin-400m
& 396M
& \textbf{61.79} & \textbf{61.98} & \textbf{62.03}
& 22.77 & 20.05 & 21.65
& \underline{34.04} & 31.61 & 33.84 \\

\bottomrule
\end{tabular}
\end{table*}

While Nash decoding turns frozen masked language models into zero-shot text generators without additional training or fine-tuning, the model performance varies across datasets, showing that Nash decoding remains dependent on the underlying model used for conditional density estimation. Better estimation leads to better performance.

Remarkably, the results in Table~\ref{tab:main_results} must not be compared directly with dataset leaderboards, which typically include models fine-tuned on task-specific training data. For example, fine-tuned models on CoQA reach $90.7$ F1~\citep{ju2019technical}. In contrast, every model in Table~\ref{tab:main_results} is evaluated zero-shot. Our comparison therefore isolates decoding methods with pretrained models under a shared evaluation protocol rather than targeting state-of-the-art performance.

For reference, \citet{radford2019language} report that GPT-2 XL achieves
approximately $54.8$ F1 on all CoQA questions. Our lower score reflects
the subset used in Table~\ref{tab:main_results}, which contains only questions
with answers of at least five tokens and thus excludes yes/no and other short
answers. We exclude these because, for a single-token answer, the Nash equilibrium
coincides with autoregressive decoding, leaving no room for revision.

\subsection{Ablation Analysis}
\label{subsec:ablation_analysis}

We study how F1 performance evolves over Nash decoding iterations and how initialization affects performance at equilibrium.  Additional ablations are provided in Appendix~\ref{app:ablation_study}.

\paragraph{Effect of Token Replacements.}
\label{sec:clapnq-dynamics}

\begin{wrapfigure}{r}{0.43\textwidth}
    \centering
    \vspace{-13pt}
    \includegraphics[width=0.95\linewidth]{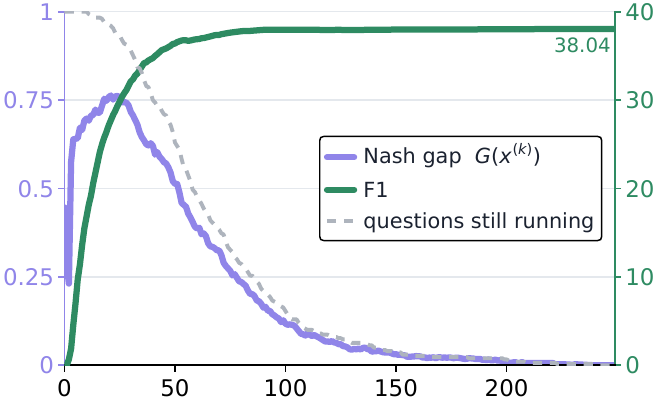}
    \caption{\small
    F1 score and Nash gap during Nash decoding iterations on CLAPNQ.}
    \label{fig:clapnq-dynamics}
    \vspace{-15pt}
\end{wrapfigure}

Figure~\ref{fig:clapnq-dynamics} tracks F1 score and the
Nash gap throughout Nash decoding with ModernBERT-Large on CLAPNQ. 
F1 measures movement toward the reference answer, whereas the Nash
gap measures the remaining incentive for any player to change its
token. Generation starts from an entirely masked sequence, whereas
ModernBERT was pretrained with a $30\%$ masking
rate~\citep{warner2025modernbert}. With short initial context, the
model is uncertain, but as early tokens
are filled, the emerging context sharpens the context
length and the maximum gap can increase. Once sufficient
context is placed, Nash updates increasingly resolve
unstable positions and drive the Nash gap toward zero. F1 continues
to increase throughout the trajectory, but more slowly in the
second half as most examples reach equilibrium,
leaving only a smaller subset with profitable token deviations (results for other datasets are available in Appendix~\ref{app:dynamics}).

\paragraph{Qualitative example of Nash equilibrium.}
Beyond aggregate scores, we inspect a representative Nash-equilibrium answer on CLAPNQ. 
For the question \emph{``where do they shoot guy's grocery games''}, Nash decoding using ModernBERT-Large fills a $48$-slot canvas and returns,

\begin{promptdisplay}
\small
\emph{``Guy's Grocery Games Season 1 was shot inside of an
actual grocery store, Field's Market in West Hills, California. For Season 2, the market
was built in a 15,500 square foot warehouse in Santa Rosa.''}
\end{promptdisplay}

It contains two well-formed sentences with proper nouns and numbers, scoring $98.5$ F1, showing that Nash equilibrium can be a coherent text even if the \emph{masked model is not fine-tuned}
(Appendix~\ref{app:clapnq-examples}).

\begin{wraptable}{r}{0.45\textwidth}
\vspace{-10pt}
\centering
\setlength{\tabcolsep}{2.5pt}
\small
\caption{\small Nash Equilibria Basins on PubMedQA.}
\label{tab:initialization}
\begin{tabular}{llccc}
\toprule
\textbf{Model} & \textbf{Size} & \textbf{F1} & \textbf{R-L} & \textbf{R-Ls} \\
\midrule

\rowcolor{SectionGray}
\multicolumn{5}{l}{\textbf{Left-to-right initialization}} \\
\rowcolor{white}
mmBERT-base & 308M & 4.22 & 6.11 & 5.38  \\
\rowcolor{LightGray}
RoBERTa-Large & 355M & 22.23 & 19.79 & 20.77  \\
ModernBERT-Large & 395M & 21.91 & 19.88 & 19.99  \\
\rowcolor{LightGray}
Ettin-400m & 396M & 17.41 & 17.29 & 17.36  \\
\midrule
\rowcolor{SectionGray}
\multicolumn{5}{l}{\textbf{All-mask initialization}} \\
\rowcolor{white}
mmBERT-base & 308M & 3.89 & 5.30 & 4.59  \\
\rowcolor{LightGray}
RoBERTa-Large & 355M & \textbf{24.99} & \textbf{21.17} & \textbf{22.69}  \\
ModernBERT-Large & 395M & 24.28 & 20.33 & 21.36  \\
\rowcolor{LightGray}
Ettin-400m & 396M & 22.77 & 20.05 & 21.65  \\

\bottomrule
\end{tabular}
\vspace{-8pt}
\end{wraptable}

\paragraph{Equilibrium Multiplicity.}
We now show empirically that multiple Nash equilibria exist. To reach different equilibria, we run Nash decoding from different initializations. Table~\ref{tab:initialization} compares two of them. In \emph{left-to-right} initialization, tokens are unmasked sequentially from left to right and are not revisited until the entire sequence has been unmasked. In contrast, \emph{all-mask} initialization starts from a fully masked answer and lets Nash decoding update any position, including revising previously unmasked tokens before all positions are filled. The two schemes therefore follow different early decoding trajectories, after which both proceed to equilibrium. The resulting differences in final performance indicate that the two trajectories can converge to different equilibria (see Appendix~\ref{app:initialization_basins} for more details). Although the performance of Nash decoding depends on initialization, we find that all-mask initialization is often effective, as shown in Table~\ref{tab:main_results}.

\subsection{From Diffusion Updates to Equilibrium Updates}
\label{sec:diffusion-backbone}

Masked diffusion models generate text through iterative full-context refinement, typically updating multiple masked positions in parallel.
The closest existing method to Nash decoding is the particularly strong single-token masked diffusion model studied in dLLM~\citep{zhou2026dllm}. dLLM provides a powerful instruction-tuned masked model, \texttt{ModernBERT-chat}, together with iterative decoding. We analyze the performance of Nash decoding using the same underlying model from dLLM.  Both decoding methods start from an all-\texttt{[MASK]} sequence and update one position at a time; dLLM selects masked positions using uniform or confidence-based ordering, whereas Nash decoding selects the position with the largest Nash gap and can revisit previously generated tokens. Because both methods can use the same masked language model, this provides a controlled comparison of the position-selection and revision rules.

\begin{table}[h]
\centering
\setlength{\tabcolsep}{5pt}
\small
\caption{
\small Nash decoding versus single-token masked-diffusion decoding using the same masked model. All rows use the instruction-tuned \texttt{ModernBERT-chat} with its corresponding prompt; details in Appendix~\ref{app:diffusion-backbone}.    
}
\label{tab:order-all}
\begin{tabular}{lccc ccc ccc}
\toprule
 & \multicolumn{3}{c}{\textbf{CoQA} }
 & \multicolumn{3}{c}{\textbf{PubMedQA} }
 & \multicolumn{3}{c}{\textbf{CLAPNQ} } \\
\cmidrule(lr){2-4}\cmidrule(lr){5-7}\cmidrule(lr){8-10}
\textbf{Decoding rule} & F1 & R-L & R-Ls & F1 & R-L & R-Ls & F1 & R-L & R-Ls \\
\midrule
Uniform-ordered & 38.65 & 40.05 & 40.08 & 30.50 & 23.51 & 27.14 & 56.15 & 50.08 & 55.00 \\
\rowcolor{LightGray}
Confidence-ordered & 41.85 & 42.21 & 42.28 & 31.74 & 24.64 & 28.50 & 58.84 & 54.30 & 58.11 \\
Max Nash gap & \textbf{43.07} & \textbf{43.22} & \textbf{43.31} & \textbf{32.11} & \textbf{24.80} & \textbf{28.75} & \textbf{59.16} & \textbf{54.67} & \textbf{58.37} \\
\bottomrule
\end{tabular}
\end{table}

Table~\ref{tab:order-all} compares Nash decoding with the two decoding rules used by dLLM while holding the model, initialization, prompt, and generation budget fixed. All methods use the released \texttt{ModernBERT-chat} checkpoint, which is based on ModernBERT-Large and supervised fine-tuned by dLLM for masked generation. dLLM's \emph{uniform} rule selects a masked position at random, whereas its \emph{confidence} rule commits to the position whose predicted token has the highest conditional probability; committed tokens are not revisited. Nash decoding instead continues revising tokens until reaching equilibrium. Across all three datasets, equilibrium-based decoding with the dLLM fine-tuned model yields higher F1 and ROUGE scores, showing that diffusion masked language models achieve even stronger question-answering performance with Nash revision, at the cost of additional test-time computation (see Appendix~\ref{app:diffusion-backbone}).

\section{Discussion and Limitations}
\label{sec:discussion}

Our paper establishes the potential of game-theoretic decoding by proving that it runs in polynomial time and by demonstrating its performance on question-answering benchmarks. These results call for future work that addresses the limitations of Nash decoding.

The convergence of Nash decoding established in Theorem~\ref{thm:convergence-rate} requires conditional probabilities to be compatible with a common joint distribution. However, the conditional distributions estimated by masked language models do not necessarily satisfy this requirement, since training does not enforce compatibility conditions such as those in Equation~\ref{eq:joint_property}. A promising direction for future research is to enforce these conditions, which would provide a provable convergence guarantee to an approximate Nash equilibrium.

\begin{figure}[h]
    \centering
    \vspace{1pt}
    \includegraphics[width=0.7\linewidth]{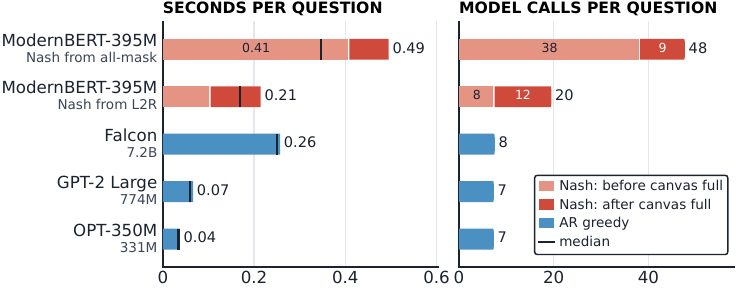}
    \caption{\small
Runtime comparison of Nash and autoregressive decoding on CoQA. The left panel reports wall-clock time per
question, and the right panel reports model calls per question. For Nash
decoding, stacked bars separate computation before the sequence is fully
unmasked from subsequent refinement. All models are evaluated on a single NVIDIA B200 GPU.
}
    \label{fig:coqa_timing}
    \vspace{1pt}
\end{figure}

A second limitation is computational cost. Although Nash decoding may require only a modest number of token replacements to reach equilibrium, each replacement is expensive with current masked architectures because computing the maximum Nash gap requires evaluating leave-one-out conditionals across all $n$ token positions. Consequently, the number of model calls per refinement step grows linearly with the generation length $n$, and the total query cost scales as $O(nK)$ for $K$ token replacements. Figure~\ref{fig:coqa_timing} illustrates this overhead on CoQA, our shortest-answer benchmark. 
The statistical gains of Nash decoding currently come at substantially higher test-time computation, and the gap becomes more pronounced for longer text. This overhead is primarily architectural rather than inherent to the equilibrium objective: masked models that can evaluate conditional probabilities for multiple positions in parallel, without a need for masking, can substantially reduce the cost (see Appendix~\ref{app:timing}).

\bibliography{iclr2027_conference}
\bibliographystyle{iclr2027_conference}

\clearpage
\appendix

\section*{Technical appendices and supplementary material}

\section{The Motivating Experiment and The Illustrated Token Game}
\label{app:spain-token-game}

\paragraph{Setup.}
All numbers in this section are produced with GPT-2 Small (124M) in
float32 with TF32 matrix shortcuts disabled, using greedy
decoding. Generation and refinement use the \emph{same} prompt.
Conditionals are computed exactly over the full $50{,}257$-token
vocabulary, and refinement replaces, at each step, the single position
with the largest gap, halting when the Nash gap falls below $10^{-9}$.

\paragraph{First example:}
The example illustrated in the introduction concerns the prompt

\begin{promptbox}[Prompt]
What famous city is the capital of Spain?
\end{promptbox}
\begin{promptbox}[Greedy continuation]
The city of Barcelona is the capital of Spain.
\end{promptbox}

Its ten tokens are the ten players; positions are numbered from~1.

\begin{center}
\begin{tabular}{llllllllll}
\texttt{The} & \texttt{ city} & \texttt{ of} & \texttt{ Barcelona} &
\texttt{ is} & \texttt{ the} & \texttt{ capital} & \texttt{ of} &
\texttt{ Spain} & \texttt{.}\\
\hspace{2mm} 1 & \hspace{5mm}2 & \hspace{4mm}3 & \hspace{10mm}4 & \hspace{4mm}5 & \hspace{4.5mm}6 & \hspace{8.5mm}7 & \hspace{4mm}8 & \hspace{7mm}9 & 10
\end{tabular}
\end{center}

Greedy decoding commits to the factually incorrect city at player~4.
At the moment that token was emitted, the decoder had seen only
``\texttt{The city of},'' and under that left context alone it preferred
``Barcelona'' to ``Madrid'' by a factor of $1.15$. Because an
autoregressive decoder conditions only on $x_{<i}$, it cannot revisit the
choice once the rest of the sentence is written. Conditioning instead on
\emph{all} other tokens, including the six to the right, reverses the
preference:

\begin{center}
\begin{tabular}{lrr}
\toprule
& $p(x_4 \mid x_{<4})$ & $p(x_4 \mid x_{-4})$\\
token & autoregressive & full context\\
\midrule
\texttt{ Madrid}    & $0.0904$          & $\mathbf{0.3343}$\\
\texttt{ Barcelona} & $\mathbf{0.1036}$ & $0.2655$\\
\texttt{ Valencia}  & $0.0519$          & $0.1507$\\
\texttt{ Spain}     & $0.0160$          & $0.0666$\\
\texttt{ Catalonia} & $0.0116$          & $0.0305$\\
\bottomrule
\end{tabular}
\end{center}

The best response is ``Madrid,'' giving $g_4 = 0.0688$. Every one of the
other nine players already plays its best response, so $G(x) = g_4$ and
player~4 is the unique improvable position. Replacing it yields

\begin{promptbox}[Nash equilibrium]
The city of \textbf{Madrid} is the capital of Spain.
\end{promptbox}

and the next scan reports zero gap at all ten positions. The sequence
log-probabilities under the prompt are $-20.2943$ before and $-20.0639$
after, a ratio of $1.2591\times$.

\paragraph{Another example.} The effect is not confined to that one prompt. A longer continuation, from
a different phrasing of the question, produces a game that takes two steps
and in which the players visibly influence one another.

\begin{promptbox}[Prompt]
What city is the political capital of Spain?
\end{promptbox}
\begin{promptbox}[Greedy continuation]
The answer is Barcelona. The city is the capital of Spain. It is the capital of Spain.
\end{promptbox}

Here all twenty tokens participate, with the erroneous city at player~4
and the pronoun ``It'' opening the third sentence at player~14. The
first scan reports

\begin{center}
\begin{tabular}{llrrr}
\toprule
player & token & $p(\text{current})$ & best response & gap\\
\midrule
4  & \texttt{ Barcelona} & $0.3039$ & \texttt{ Madrid} $(0.3978)$ & $0.0938$\\
14 & \texttt{ It}        & $0.2709$ & \texttt{ Madrid} $(0.3144)$ & $0.0435$\\
\bottomrule
\end{tabular}
\end{center}

with zero gap at the remaining eighteen positions, so $G(x) = 0.0938$ and
player~4 moves first:

\begin{promptbox}[After the first replacement]
The answer is \textbf{Madrid}. The city is the capital of Spain. It is the capital of Spain.
\end{promptbox}

This is where the coupling between players shows itself. Correcting
player~4 does not leave player~14 untouched: its gap more than doubles,
from $0.0435$ to $0.0972$ ($p(\texttt{ It}) = 0.3160$ against
$p(\texttt{ Madrid}) = 0.4132$), because the repaired first sentence makes
the proper noun a better subject for the third. Player~14 therefore moves
second:

\begin{promptbox}[Nash equilibrium]
The answer is Madrid. The city is the capital of Spain. \textbf{Madrid} is the capital of Spain.
\end{promptbox}

A final scan reports a gap of exactly zero at all twenty positions, so the
text is a Nash equilibrium, and the algorithm halts. The sequence
log-probabilities are $-25.8889$ for the greedy text, $-25.6198$ after the
first replacement and $-25.3516$ at equilibrium: each step raises the
likelihood, by $1.309\times$ and $1.308\times$ respectively, for a total of
$1.711\times$.

\section{Proof of the Autoregressive--Nash Separation}
\label{app:proof-ar-nash-separation}

\paragraph{Lemma description.}
Lemma~\ref{lem:ar-nash-separation} shows that greedy autoregressive decoding and Nash equilibrium can differ substantially even when both are defined by the same joint distribution. Specifically, for every even sequence length $n$, there exists a strictly positive autoregressive distribution for which the greedy output is not an equilibrium, with Nash gap greater than $0.30$. The unique Nash equilibrium differs from the greedy output at exactly half of the token positions and has exponentially larger joint probability:
\[
\frac{p(x^{\mathrm{Nash}})}{p(x^{\mathrm{AR}})}
=
\exp(\gamma n),
\qquad
\gamma
=
\frac{1}{2}\log\!\left(\frac{539}{289}\right)
\approx 0.31164.
\]
Thus, the lemma establishes that locally optimal left-to-right decisions can accumulate into a sequence that is both far from equilibrium and exponentially less probable than a full-context equilibrium.

\paragraph{Proof.}
Consider a block $(a,b)\in\{0,1\}^2$ with
$p(a,b)=p_1(a)p_2(b\mid a)$, where
$p_1(0)=0.51$, $p_1(1)=0.49$,
$p_2(1\mid0)=0.51$, and $p_2(1\mid1)=0.99$.
Greedy decoding selects $(a,b)=(0,1)$. However,
$p(0,1)=0.2601$ and $p(1,1)=0.4851$, so given $b=1$,
the full-conditional best response is $a=1$. Hence, 
\[
G(0,1)\ge\frac{(0.4851-0.2601)}{(0.4851+0.2601)}
=125/414>0.30.
\]
Moreover, $b=1$ is the strict best response for either value of $a$, while
$a=1$ is the strict best response given $b=1$. Thus, $(1,1)$ is the unique
Nash equilibrium of the block.

Now take $m=n/2$ independent copies of this block. Greedy decoding produces
$x^{\mathrm{AR}}=(0,1,\ldots,0,1)$, whereas the unique Nash equilibrium is
$x^{\mathrm{Nash}}=(1,1,\ldots,1,1)$. Therefore,
$d_{\mathrm H}(x^{\mathrm{AR}},x^{\mathrm{Nash}})=n/2$, and
\[
\frac{p(x^{\mathrm{Nash}})}{p(x^{\mathrm{AR}})}
=
\left(\frac{0.49\cdot0.99}{0.51\cdot0.51}\right)^{n/2}
=
\left(\frac{539}{289}\right)^{n/2}
=
\exp(\gamma n)
\approx
\exp(0.31164\,n).
\]

\section{Equilibrium Dynamics on WikiText-103}
\label{app:wikitext}

Section~\ref{sec:empirical-dynamics} replaces the exact conditionals assumed by
Theorem~\ref{thm:convergence-rate} with estimates from a pretrained masked model and asks
whether the dynamics still behave as the analysis predicts. This appendix specifies that
experiment and reports the measurements behind Figure~\ref{fig:theory-validation}.

\subsection{Setup}

What is
tested is the convergence rate and likelihood increase in Theorem~\ref{thm:convergence-rate}, not the question-answering performance. We take
$500$ prompts of exactly $512$ tokens from the WikiText-103 \emph{test} split
\citep{merity2017pointer} and generate a $64$-token continuation of each with
\texttt{ModernBERT-Large}. To estimate the likelihood, we used an external autoregressive referee rather
than the model likelihood estimate. Using an external model lets us evaluate the likelihood without overfitting to our own model's estimates. 

Numerics follow Appendix~\ref{app:config}: greedy selection throughout, \texttt{float32}
with TF32 disabled and verified against a \texttt{float64} reference, the SDPA attention
implementation, and a $1024$-token context. The action set is the full vocabulary minus
the four non-mask special IDs, leaving $50{,}364$ actions. \texttt{[MASK]} is one of
them. Keeping it legal is what makes the token gap of Equation~\ref{eq:token-gap} well
defined at a position that has not yet been written, and it allows a player to return its
position to the masked state if that is its best response; no player ever does in practice.

\subsection{Procedure}

\emph{Nash decoding} runs Algorithm~\ref{alg:nash-decoding} with tolerance
$\varepsilon=0$. At each iteration, the algorithm computes the token gap at all $64$
positions, writes or revises the largest-gap position with its best response, and stops when
$G(x)\leq 10^{-4}$, which certifies an exact equilibrium. Figure~\ref{fig:theory-validation}
measures the trajectory only after all
positions unmasked; we index this initial fully instantiated sequence as $k=1$, and
subsequent iterations correspond to revisions of that sequence. We begin measurement
at this point because sequence-level quantities such as likelihood and perplexity are
not meaningfully comparable while the sequence still contains \texttt{[MASK]} positions.

\subsection{Convergence and Cycle Termination}

Table~\ref{tab:app-wikitext} summarizes refinement on the
$500$ WikiText-103 continuations. Refinement performs $1{,}300$
token replacements in total, with a median of two per continuation.
Of the $500$ completed initial sequences, $118$ already satisfy the
equilibrium criterion before any refinement. The final sequence
differs from its initialization at an average of $2.29$ of the
$64$ token positions.

\paragraph{Equilibrium and cycle termination.}
We distinguish reaching an equilibrium from stopping at a repeated
state. An equilibrium is certified when a full scan of all positions
finds $G(x)=0$: no unilateral token replacement improves the
position's conditional probability under the masked model.
This certificate concerns the estimated conditionals and the
numerical precision used in the experiment; it does not establish
global optimality, factual correctness, or compatibility with a
joint distribution.

For cycle detection, we retain the complete token sequence of every
visited state, including the initial sequence. After each token
replacement, we check whether the resulting sequence has appeared
previously. If so, refinement stops and returns that repeated
sequence, with its termination status recorded as a cycle.
The returned sequence remains in the reported statistics and
likelihood diagnostics; it is not counted as an equilibrium.
Here, a cycle refers to recurrence of the complete sequence across
updates, not merely to repeated words or phrases within a
continuation.

While Theorem~\ref{thm:convergence-rate} rules out cycles, they may still arise when the true conditional probabilities are replaced by a transformer's estimates, as these estimates may not be compatible with a common joint
distribution (Section~\ref{sec:nash-decoding}). A best-response
update improves the conditional probability at the selected
position but can change the incentives at other positions.
Subsequent updates can therefore return the sequence to an earlier
state. Thus, we terminate the revision process once a cycle emerges.

\begin{table}[h]
\centering
\small
\caption{
    \textbf{Refinement on $500$ WikiText-103 continuations} with
    \texttt{ModernBERT-Large}, counted from the completed sequence $x^{(1)}$.
    Construction is a separate max-gap order pass of $64$ model calls.
}
\label{tab:app-wikitext}
\begin{tabular}{lr}
\toprule
\textbf{Quantity} & \textbf{Value} \\
\midrule
Prompts, tokens per prompt, continuation slots $T$ & $500$, $512$, $64$ \\
Action set (vocabulary minus four special IDs) & $50{,}364$ \\
\midrule
Trajectories reaching an exact equilibrium & $494$ ($98.8\%$) \\
Trajectories stopped at a repeated state & $6$ ($1.2\%$) \\
Already at equilibrium when construction ends & $118$ ($23.6\%$) \\
\midrule
Refinement updates per text (mean / median / max) & $2.60$ / $2$ / $36$ \\
Tokens changed by refinement (mean) & $2.29$ \\
Updates that write \texttt{[MASK]} & $0$ of $1{,}300$ \\
\midrule
Model calls per text (construction $+$ refinement) & $64+230$ \\
Seconds per text & $3.3$ \\
\bottomrule
\end{tabular}
\end{table}

\subsection{Equilibrium Test}
\label{app:equilibrium_diagnostic}
Figure~\ref{fig:nash-gap-bound} plots $\ln k$ against the natural logarithm of the mean
over the $500$ texts of the running minimum $\min_{j\leq k}G(x^{(j)})$. On these axes, a
$1/k$ rate is a straight line of slope $-1$, so the reference curve fixes the rate implied
by Theorem~\ref{thm:convergence-rate} and not its constant. The empirical curve falls
from $0.342$ at $k=1$ to $6.9\times10^{-5}$ at $k=36$, and stays below the reference
throughout.

We also examine a numerical analogue of the bound using ModernBERT's pseudo-likelihood. Define
\[
\tilde p_0=\prod_i q_\theta(x_i^{(1)}\mid x_{-i}^{(1)};\mathrm{prompt}),
\]
where $\tilde p_0$ is a product of full-context conditionals rather than a joint sequence probability. We compare the observed running minimum with the proxy $2\log(1/\tilde p_0)/k$, which has the same form as the bound in Equation~\ref{eq:convergence-rate}. The empirical running minimum remains below this proxy for all $500$ texts and all $k$, with a median ratio of $71\times$; the mean value of $2\log(1/\tilde p_0)$ is $21.4$. Because ModernBERT's conditionals need not be induced by a common joint distribution, this comparison is only an empirical test.


\subsection{Where the Guarantee Does Not Transfer}

Equation~\ref{eq:monotone-improvement} requires the sequence probability to rise by a
factor of at least $1+G(x^{(k)})$ at every update. Substituting the masked model's
pseudo-likelihood for $p$, that inequality holds on $71.5\%$ of updates, and the
pseudo-likelihood increases on $75.2\%$.

These failures are the expected consequence of Section~\ref{sec:nash-decoding}. An update
raises the estimated conditional at the position it rewrites, but it also perturbs the
conditionals at every other position, and without a common joint distribution nothing
relates the two. Monotone improvement is a property of the idealized game rather than of the only relying transformers estimates. However, the Nash gap does not suffer from this issue, since it depends only on conditional probabilities rather than the joint probability.


\providecolor{LightGray}{HTML}{F5F7F9}
\providecolor{SectionGray}{HTML}{E9EDF2}

\section{The Three Question-Answering Benchmarks}
\label{app:datasets}

Section~\ref{sec:experiments} evaluates Nash decoding on CLAPNQ,
PubMedQA, and CoQA. This appendix explains the choice of benchmarks,
specifies the evaluated subsets and decoding budgets, and describes
their differences in sequence length and lexical overlap with the
supplied context.

\subsection{Why These Three Benchmarks}
\label{app:data-why}

We choose these benchmarks to compare Nash decoding across three datasets for
conversational text analysis, long-answer generation, and scientific
analysis. CoQA requires short answers
used in prior studies~\citep{reddy-etal-2019-coqa,dong2019unified,xiao2020erniegen}.
CLAPNQ evaluates the quality of long-answers: answers often combine
information from non-contiguous parts of a passage~\citep{rosenthal-etal-2025-clapnq,kim2025accelerating}. For example, asked where the Narmada river is located, the reference answer
joins the passage's first sentence (``a river in central India'') with its
fourth (``rises from Amarkantak Plateau near Anuppur district''), skipping the
two sentences in between.
PubMedQA extends the comparison to a technical domain, where our
target is a biomedical conclusion withheld from the input
abstract~\citep{jin-etal-2019-pubmedqa}. Prior evaluations on PubMedQA
include autoregressive models and diffusion models, under
their respective task formulations~\citep{luo2022biogpt,huang-etal-2023-diffusion}.

The benchmarks differ in both the amount of text to generate and
how that text relates to the context. In our evaluated subsets,
CoQA questions average $6.6$ ModernBERT tokens and often depend
on conversational history, whereas CLAPNQ answers average $63.8$
tokens and require cohesive composition from passage material.
PubMedQA adds scientific evidence and specialized terminology,
with substantially less answer wording available directly in the
context. Together, these settings let us examine whether the gains from Nash decoding persist across diverse question-answering scenarios.

\subsection{Preprocessing}
\label{app:data-selection}

\paragraph{CLAPNQ.}
CLAPNQ~\citep{rosenthal-etal-2025-clapnq} pairs a question from
Natural Questions with a single gold passage and a cohesive,
long-form answer. Because its test split is not publicly available,
we use the $300$ answerable development questions. We exclude the
$300$ unanswerable questions because the evaluated decoding
protocol does not include abstention; the dataset paper also
reports answerable-only evaluations. The retained questions cover
$297$ distinct passages and have $485$ reference answers:
$160$ questions have one reference, $131$ have two, and $9$ have
seven, giving a mean of $1.62$ references per question.
Multi-reference evaluation takes the best-reference score.

\paragraph{PubMedQA.}
PubMedQA~\citep{jin-etal-2019-pubmedqa} pairs title-style biomedical
research questions with scientific abstracts. We use the official
$500$-question test half of PQA-L, reconstructed from the released
seed-0 two-fold split and checked against the expected split size.
Each question has a distinct abstract and one reference answer:
the abstract's conclusion, which is excluded from the input context. 
The corpus additionally labels each
question yes / no / maybe ($276$ / $169$ / $55$ here), but we never use the
label as a target: the task in our setting is to generate the long conclusion,
not to classify.

\paragraph{CoQA.}
CoQA~\citep{reddy-etal-2019-coqa} contains conversations
about passages, with questions about the conversations.
The development split contains $7{,}983$ questions over $500$
passages. We retain questions whose gold answer contains at least
five ModernBERT tokens, yielding $1{,}814$ questions from $460$
conversations and $459$ distinct passages. The retained questions
span five domains: Wikipedia ($447$), CNN ($400$), RACE ($360$),
Project Gutenberg ($351$), and MCTest ($256$). Dialogue history is
included in the prompt. The rationale for the length filter is
given in the next subsection; results on the unfiltered development set, all
$7{,}983$ questions, are reported in Table~\ref{tab:coqa_results}.

\subsection{CoQA Filtering by Answer Length}
\label{app:coqa-filter}

Our evaluation focuses on interactions among token players during
Nash decoding. Nearly half of the gold answers in the full CoQA
development split are a single word, and $18.4\%$ are bare
\emph{yes} or \emph{no} responses. Such answers often occupy only
one or two tokens. For single-token answers, the game has
one player: its best response depends on the fixed prompt, with
no other generated tokens to influence or respond to. These cases
therefore provide limited scope for studying the interactions
that motivate Nash decoding.

Requiring at least five ModernBERT tokens focuses the evaluation
on answers with multiple token players, whose conditional
preferences can change as other players replace their tokens.
This lets us study coordination and refinement within an answer.
The threshold is an experimental design choice, not an algorithmic
requirement, and answer quality still depends on the underlying
model's capabilities. All models are evaluated on the same retained
questions; the reported CoQA results apply to this filtered subset.

\subsection{Tokenization and Reference Processing}
\label{app:data-protocol}

\paragraph{Tokenization.}
Descriptive token counts use
\texttt{answerdotai/ModernBERT-Large}~\citep{warner2025modernbert}, except where a different tokenizer
is explicitly identified. Answers are tokenized with a leading
space, as \texttt{" " + target}, matching the dataset builder.
Word counts are whitespace-delimited. The decoding budget is
computed separately in each evaluated model's tokenizer, as
described in Appendix~\ref{app:data-scale}.

\paragraph{Normalization and reference handling.}
Lexical statistics use the evaluation normalizer: lowercasing,
removing ASCII punctuation and the articles \emph{a}, \emph{an},
and \emph{the}, and collapsing whitespace. Copying and overlap
therefore refer to normalized text, consistent with the F1 metric.
When an example has multiple reference answers, we compute each lexical statistic against every reference and report the highest value, matching the best-reference convention used for F1 evaluation.

\subsection{Scale, Length, and the Decoding Budget}
\label{app:data-scale}

Table~\ref{tab:app-data-scale} summarizes the exact evaluated
questions. The benchmarks differ by almost an order of magnitude
in mean answer length. CoQA has the shortest questions and answers
but the longest prompts because its input includes dialogue history.

For each example, the reference answer determines an oracle
continuation-length budget in the evaluated model's own tokenizer.
Masked generators fill that many continuation positions, and
autoregressive baselines generate that many tokens with
end-of-sequence generation disabled. ModernBERT counts provide
a common basis for describing dataset lengths; the GPT-2 BPE row
illustrates how the same answers yield different budgets under
another tokenizer.

\begin{table}[h]
\centering
\small
\resizebox{0.8\linewidth}{!}{%
\begin{tabular}{lccc}
\toprule
 & \textbf{CLAPNQ} & \textbf{PubMedQA} & \textbf{CoQA}$^{\dagger}$ \\
\midrule
Split
 & dev, answerable
 & PQA-L test (seed 0)
 & dev, gold $\geq5$ tok. \\
Questions scored
 & $300$ & $500$ & $1{,}814$ \\
Distinct contexts
 & $297$ & $500$ & $459$ ($460$ conv.) \\
References per question
 & $1.62$ & $1$ & $4$ ($2.36$ distinct) \\
\midrule
\multicolumn{4}{l}{
    \emph{Mean length in ModernBERT tokens (words in parentheses)}
} \\
Context
 & $195.3$ ($170.8$)
 & $287.0$ ($202.3$)
 & $355.2$ ($267.0$) \\
Question
 & $10.4$ ($9.1$)
 & $17.8$ ($13.1$)
 & $6.6$ ($5.3$) \\
Dialogue history $+$ question
 & --- & --- & $140.7$ \\
Full prompt
 & $222.7$ & $319.7$ & $495.9$ \\
\midrule
\multicolumn{4}{l}{\emph{Reference answer length}} \\
Mean ModernBERT tokens (words)
 & $\mathbf{63.8}$ ($51.7$)
 & $\mathbf{48.3}$ ($39.1$)
 & $\mathbf{7.5}$ ($5.6$) \\
Median / 90th percentile / maximum
 & $56/107/231$
 & $45/77/154$
 & $7/11/40$ \\
Mean GPT-2 BPE tokens
 & $63.1$ & $50.2$ & $7.4$ \\
Mean sentences per answer
 & $2.26$ & $1.91$ & $1.01$ \\
\bottomrule
\end{tabular}%
}
\caption{
    \textbf{Scale and length of the evaluated subsets.}
    Token lengths use ModernBERT-Large unless otherwise stated;
    parenthesized word counts are whitespace-delimited. Reference
    counts are averages per question, with distinct references counted
    after normalization. Oracle decoding budgets use the reference
    answer length in each evaluated model's tokenizer.
    $^{\dagger}$CoQA includes only development questions whose gold
    answer has at least five ModernBERT tokens
    (Appendix~\ref{app:coqa-filter}).
}
\label{tab:app-data-scale}
\end{table}

\subsection{Lexical Overlap and Answer Composition}
\label{app:data-grounding}

Table~\ref{tab:app-data-grounding} characterizes how reference
answers reuse the supplied context. Token and content-word overlap
measure which answer words occur in the context; novel bigrams
measure word pairs absent from it. The contiguous-span and
longest-run statistics describe how this shared wording is arranged.
All comparisons use the normalization specified in
Appendix~\ref{app:data-protocol}.

The sentence-selection diagnostic greedily selects context
sentences to cover distinct answer content words, with a target
coverage of $90\%$. It records the number of selected sentences
and the distance, in sentence positions, between the first and
last selected sentences. We also score two copying baselines:
the entire context and the single context sentence with the
highest reference F1. The latter uses the reference to select
the sentence. These scores describe the potential of these
particular copying strategies at their natural output lengths;
they do not impose the oracle length budgets used for model
comparisons or bound all possible copying strategies.


\begin{table}[h]
\centering
\small
\resizebox{0.8\linewidth}{!}{%
\begin{tabular}{lccc}
\toprule
 & \textbf{CLAPNQ} & \textbf{PubMedQA} & \textbf{CoQA}$^{\dagger}$ \\
\midrule
\multicolumn{4}{l}{\emph{Lexical overlap with the context}} \\
Gold tokens occurring in the context
 & $97.6\%$ & $\mathbf{60.0\%}$ & $99.1\%$ \\
Gold content words occurring in the context
 & $97.4\%$ & $\mathbf{49.3\%}$ & $98.9\%$ \\
Novel bigrams (absent from the context)
 & $10.9\%$ & $\mathbf{76.8\%}$ & $6.5\%$ \\
Answers with $<50\%$ content-word overlap
 & $0.3\%$ & $\mathbf{52.0\%}$ & $0.6\%$ \\
\midrule
\multicolumn{4}{l}{\emph{Arrangement of shared wording}} \\
Gold answer is a contiguous span after normalization
 & $4.0\%$ & $0.0\%$ & $\mathbf{79.8\%}$ \\
Longest shared contiguous run / answer length
 & $53.7\%$ & $12.8\%$ & $\mathbf{91.9\%}$ \\
Sentences selected ($90\%$ coverage target)
 & $2.50$ & $3.39$ & $1.01$ \\
Answers with $\geq2$ selected context sentences
 & $89.3\%$ & $95.0\%$ & $2.4\%$ \\
Answers with $\geq3$ selected context sentences
 & $37.3\%$ & $72.6\%$ & $0.3\%$ \\
Distance between first and last selected sentence
 & $3.47$ & $6.19$ & $0.06$ \\
\midrule
\multicolumn{4}{l}{\emph{Copying baselines (F1)}} \\
Entire context as the answer
 & $50.7$ & $18.4$ & $4.9$ \\
Best single context sentence
 & $67.6$ & $34.7$ & $45.0$ \\
\bottomrule
\end{tabular}%
}
\caption{
    \textbf{Lexical overlap and arrangement of passage material.}
    Statistics use normalized text and the most favorable reference
    for each measure. The sentence-selection diagnostic greedily
    covers distinct answer content words with a $90\%$ target.
    Copying baselines score either the full context or its best
    single sentence against the references.
    $^{\dagger}$CoQA uses the filtered subset described in
    Appendix~\ref{app:coqa-filter}.
}
\label{tab:app-data-grounding}
\end{table}

\paragraph{CoQA: predominantly span-based answers.}
In the retained subset, approximately four in five answers are
contiguous context spans after normalization. The longest shared
run accounts for $91.9\%$ of answer length on average, the
sentence-selection diagnostic uses at most one sentence in
$97.6\%$ of cases, and only $0.9\%$ of answer tokens are absent
from the context. The task therefore largely involves selecting
a short answer span, with its length supplied by the oracle budget.
The full-context copying score of $4.9$ F1 reflects the substantial
length mismatch between a roughly $355$-token passage and a
$7.5$-token answer; this diagnostic does not measure the difficulty
of locating the appropriate span.

\paragraph{CLAPNQ: composition from passage wording.}
Although $97.6\%$ of reference tokens occur in the passage, only
$4\%$ of complete answers form a contiguous span. The longest
shared run covers just over half of an answer. The sentence-selection
diagnostic typically combines material from two or three sentences,
with the first and last selected sentences approximately three
and a half positions apart on average. CLAPNQ thus combines high
lexical overlap with the need to assemble a cohesive answer.
Copying the entire passage scores $50.7$ F1, while selecting the
best single sentence scores $67.6$ F1. These reference-informed
diagnostics show that passage reuse can receive substantial
credit under the overlap metric.

\paragraph{PubMedQA: greater lexical novelty.}
Two fifths of reference tokens and more than three quarters of
reference bigrams are absent from the abstract. About half of
the answers have fewer than half their content words in the
context, and none is a complete contiguous context span. The
reported content-word overlap is only $49.3\%$, so selecting
additional sentences cannot recover words that are absent from
the entire abstract. In contrast, $70.9\%$ of question tokens
occur in the abstract, the highest proportion among the three
benchmarks. The question therefore has substantial lexical
overlap with its context even when the reference conclusion
requires considerable new wording. This setting places greater
emphasis on generating a technical answer from scientific evidence
than on recovering a passage span.

\section{Experimental configuration}
\label{app:config}

This appendix specifies the prompts, reference-derived output lengths, token restrictions,
decoding procedures, stopping rules, and metrics used for the comparisons in
Table~\ref{tab:main_results}. Rows differ in two respects: the model and the decoding
procedure (autoregressive, one-shot masked model, or Nash gap decoding).
Everything else, such as answer-length budget, token restrictions, precision, context
window, and metric are the same across decoding methods.

\subsection{Prompt}

Two prompt formats are used in the experiments. CLAPNQ and PubMedQA use a soft-instruction prompt: the
evidence text, the question, a one-line instruction, and the token \texttt{Answer:} on
which generation begins.

\begin{promptdisplay}
\small
\vspace{2pt}
\texttt{\{passage\}}

\vspace{5pt}
\textbf{Question:} \texttt{\{question\}}

\vspace{5pt}
This question is answered completely with evidence from the passage.

\vspace{4pt}
\textbf{Answer:}
\vspace{2pt}
\end{promptdisplay}

\noindent

The instruction line is byte-identical across every item of a dataset and across every
system. PubMedQA differs in one word, \emph{abstract} instead of \emph{passage}, because
its evidence is an abstract.

For CoQA, we follow the prompting format of~\citet{radford2019language}: the passage is followed by the dialogue history and the current question, using \texttt{Q:} and \texttt{A:} delimiters without an additional instruction.

\begin{promptdisplay}
\small
\vspace{2pt}
\texttt{\{passage}

\vspace{5pt}
\textbf{Q:} \texttt{{q\textsubscript{1}}}\\
\textbf{A:} \texttt{a\textsubscript{1}}

\vspace{3pt}
\hspace{6pt}$\vdots$

\textbf{Q:} \texttt{{q\textsubscript{t}}}\\
\textbf{A:} \texttt{a\textsubscript{t}\}}

\vspace{3pt}
\textbf{Q:} \texttt{\{question\}}
\textbf{A:}
\vspace{2pt}
\end{promptdisplay}

\noindent
For every preceding turn, the dialogue history contains the corresponding \emph{gold} answer rather than a model-generated response. Consequently, predictions from earlier turns cannot propagate errors to later turns, and all models receive the same conversational context. The masked language models receive a whitespace-normalized version of the \texttt{Q:} and \texttt{A:} prompt, where newline delimiters are replaced by spaces and a leading space is added. We therefore evaluate each question independently.

No system receives a chat template, a system message, or few-shot examples. The
initialization ablation uses the same dataset-specific prompt defined above and changes only the initialization procedure.

\subsection{Text length}

Length is not a free parameter. Each system is given an \emph{oracle} budget
$B=\lvert\mathrm{tok}(\text{`\,'}+a^{\star})\rvert$: the number of tokens the gold answer
$a^{\star}$, prefixed with a single space, occupies in that system's own tokenizer. An
autoregressive model emits exactly $B$ tokens; a masked model is given a canvas of exactly
$T=B$ slots. Budgets are therefore matched in content rather than in count, and systems
that share a tokenizer receive byte-identical budgets. 

The budget is computed per example, never averaged or rounded. Mean budgets run from
$7.3$ to $8.5$ tokens on CoQA, $47.6$ to $59.2$ on PubMedQA, and $63.1$ to $73.9$ on
CLAPNQ, the spread being entirely a matter of tokenizer. If a reference tokenizes to zero tokens in some vocabulary, the budget is
zero and that system emits the empty string; this does not occur for any system in
Table~\ref{tab:main_results}.

\subsection{Banned tokens}

The action set excludes \emph{structural} tokens only: each tokenizer's
\texttt{all\_special\_ids}, together with the untrained embedding slots
$[\lvert\mathrm{tok}\rvert,\ \texttt{vocab\_size})$ that exist when a checkpoint pads its
embedding matrix past the end of its vocabulary. No content word, scaffold label, or
answer-like string is ever banned.

The size of this set is a property of the tokenizer, not a choice. Banning end-of-sequence is what makes the length
comparison exact: with EOS unavailable, an autoregressive model cannot stop early, so it
emits exactly $B$ tokens, matching the masked decoders, which fill exactly $T$ slots.

For the masked encoders, \texttt{[MASK]} is deliberately \emph{not} banned. It is a legal
action at every position, which is what makes the Nash gap well-defined from the first
step, when every position still holds a mask.

\subsection{Decoding and numerics}

Decoding is greedy everywhere: $\arg\max$ over the unbanned logits, for the
autoregressive models, the diffusion samplers, and the Nash update alike. No temperature,
top-$k$, top-$p$, or sampling is used, and no random seed influences any reported number.

All models run in \texttt{float32} with TF32 disabled on both matmul and cuDNN, and with
the SDPA attention implementation. TF32 being off is asserted rather than assumed: each
run multiplies two $2048\times2048$ matrices and requires the \texttt{float32} result to
agree with a \texttt{float64} reference to a relative error below $10^{-5}$, which fails
by three orders of magnitude if TF32 is silently active.

\subsection{Context window}

The context limit is $1024$ tokens with one margin token reserved, so a prompt may use at
most $1024-B-1$ tokens. Longer prompts are truncated from the \emph{left}, preserving the
question and the instruction, which sit at the end. CLAPNQ and PubMedQA never require
truncation; on CoQA it affects between $0.4\%$ and $1.1\%$ of items, depending on the
tokenizer. 

One exception is inherent to the checkpoint: RoBERTa-Large has
$514$ position embeddings with a padding offset of two, giving it an effective limit of
$512$ tokens.

\subsection{Nash gap decoding}

The masked encoder starts from an all-\texttt{[MASK]} canvas of $T$ slots and repeatedly
applies the update
\[
i^{\star}=\arg\max_i\Big[\max_v p(v\mid x_{-i})-p(x_i\mid x_{-i})\Big],
\]
committing the best response at $i^{\star}$ and stopping when that gap is
$\le 0$, which is an exact equilibrium. The gap is recomputed at \emph{all} $T$ positions
after every committed token, so a position that is already filled can be revised before
the canvas is complete; construction and refinement are not separate phases. The gap
tolerance is exactly $0$, and the number of updates is capped at $\max(64,10T)$.

The \emph{all-mask} rows use that rule from the all-mask state. The \emph{L2R} rows are a
baseline that differs only in the order in which positions are first filled; they are
otherwise the same decoder, the same budget, and the same bans.




\subsection{Metrics}

We report three numbers per dataset. Token F1 is the SQuAD-style score against the
\emph{best} reference, after normalization.
ROUGE-L and ROUGE-Lsum are F-measures computed with \texttt{rouge\_score} through
\texttt{score\_multi} against all references, with stemming disabled. Both are computed
on sentence-split text; the split changes only ROUGE-Lsum, since ROUGE-L tokenization
treats a newline like any other separator.

CoQA has four references per turn, CLAPNQ one, two, or seven, and PubMedQA one. For CoQA, we also computed the official leave-one-annotator-out accumulation; it is $0.4$--$1.6$ points below the best-reference number, the smaller gaps occurring for the weakest systems, and it preserves the ordering of systems.

\subsection{Reproducibility and Verification}

All decoding procedures are deterministic under the reported experimental configuration. Autoregressive baselines use greedy next-token decoding, while Nash decoding selects the position with the largest Nash gap and replaces it with its highest-probability token. We also fix the exact model checkpoint used in each experiment to ensure that the underlying model remains unchanged across runs, and selected experiments repeated on a different compute node produced identical outputs. Finally, we verify all reported results directly from the saved per-example predictions by recomputing the evaluation metrics and checking them against the values reported in all result tables, ensuring consistency between the reported numbers and the underlying experimental outputs. 

\section{Extended Question-Answering Results}
\label{app:qa-results}

Tables~\ref{tab:coqa_results}--\ref{tab:clapnq_results} provide the full question-answering comparison underlying the results in Section~\ref{sec:experiments}. Nash decoding turns frozen masked models of only $308$--$396$M parameters into zero-shot generators without additional training or fine-tuning. Despite their small size, these models outperform many substantially larger autoregressive models. For example, the $396$M Ettin model achieves the strongest Nash result on CoQA while outperforming several $6$--$8$B autoregressive models, corresponding to nearly an $18\times$ difference in parameter count. Similar behavior appears on PubMedQA and CLAPNQ, where the strongest Nash-decoded model also exceeds multiple multi-billion-parameter autoregressive baselines.

The strongest model nevertheless varies across datasets: Ettin-400M performs best on CoQA, RoBERTa-Large on PubMedQA, and ModernBERT-Large on CLAPNQ. Since the decoding procedure is unchanged, the variation in performance reflects differences in the full-context conditional probability estimates provided by the underlying masked model. Nash decoding therefore does not imply that these small masked models dominate all larger autoregressive models; stronger autoregressive models can still achieve higher task performance, and models not included in our comparison may also do so. Rather, the results show that enforcing token-level equilibrium can yield surprisingly high F1 scores from comparatively small masked models, motivating the use of Nash decoding with larger and more accurate masked models.

The \texttt{Calls} column also makes the computational tradeoff explicit. Autoregressive decoding requires approximately one forward pass per generated token, whereas Nash decoding repeatedly evaluates full-context token replacements until equilibrium is reached and therefore uses substantially more test-time computation. In contrast, masked language models provide a lower-cost non-autoregressive baseline: one-shot generation requires only a single forward pass. However, under the same evaluation protocol, these lower-compute baselines remain below the best Nash-decoding answers across all three benchmarks.

To check that the length filter does not drive these conclusions, we repeated the CoQA
comparison on the \emph{entire} development set---all $7{,}983$ turns, including the
$1{,}538$ \emph{yes}/\emph{no}/\emph{unknown} answers and every short answer the
$\geq 5$-token cohort removes---under an otherwise identical protocol. The ranking is
unchanged: the same model with Nash decoding maintains a better score. 

\begin{table*}[h]
\centering
\setlength{\tabcolsep}{3.8pt}
\small
\caption{Results on CoQA under the \textbf{GPT-2 paper prompt}. The left block scores the
1{,}814 questions whose gold answer has $\geq 5$ ModernBERT tokens; the right block scores
\textbf{all 7{,}983 CoQA development turns}, including the yes/no/unknown answers. The protocol is identical throughout and identical to
Table~\ref{tab:main_results}: oracle answer length in each model's own tokenizer,
structural-token bans only, greedy decoding, fp32. \textbf{Calls} is the mean number of
forward passes per example; for autoregressive models it equals the oracle
token budget $B$. \textbf{Char} is the mean character length of the scored answer. Calls and
Chars are reported for the $\geq 5$ cohort.}
\label{tab:coqa_results}
\begin{tabular}{llccccc|ccc}
\toprule
& & \multicolumn{5}{c|}{\textbf{Gold $\geq 5$ tokens} ($n=1{,}814$)}
  & \multicolumn{3}{c}{\textbf{All} ($n=7{,}983$)} \\
\cmidrule(lr){3-7}\cmidrule(lr){8-10}
\textbf{Model} & \textbf{Size} & \textbf{F1} & \textbf{R-L} & \textbf{R-Ls} & \textbf{Calls} & \textbf{Char}
 & \textbf{F1} & \textbf{R-L} & \textbf{R-Ls} \\

\midrule
\rowcolor{SectionGray}
\multicolumn{10}{l}{\textbf{One-shot masked LMs (not fine-tuned)}} \\
\rowcolor{white}
RoBERTa-Large~\citep{liu2019roberta} & 355M & 28.00 & 31.45 & 31.46 & 1.0 & 21 & 47.46 & 49.29 & 49.29 \\
\rowcolor{LightGray}
ModernBERT-Large~\citep{warner2025modernbert} & 395M & 39.54 & 42.92 & 42.93 & 1.0 & 21 & 58.42 & 60.17 & 60.17 \\
Ettin-400m~\citep{weller2025ettin} & 396M & 43.87 & 46.22 & 46.24 & 1.0 & 21 & 66.00 & 66.83 & 66.83 \\

\midrule
\rowcolor{SectionGray}
\multicolumn{10}{l}{\textbf{Autoregressive decoders}} \\
\rowcolor{white}
GPT-2 Small~\citep{radford2019language} & 124M & 21.20 & 24.50 & 24.55 & 7.4 & 24 & 26.03 & 28.64 & 28.65 \\
\rowcolor{LightGray}
SmolLM2-135M~\citep{allal2025smollm2} & 135M & 32.59 & 34.92 & 34.95 & 7.6 & 25 & 40.88 & 42.66 & 42.67 \\
Pythia-160M~\citep{biderman2023pythia} & 160M & 15.15 & 18.87 & 18.89 & 7.5 & 24 & 19.33 & 22.18 & 22.18 \\
\rowcolor{LightGray}
OPT-350M~\citep{zhang2022opt} & 331M & 27.16 & 29.86 & 29.94 & 7.4 & 25 & 31.52 & 33.69 & 33.71 \\
GPT-2 Medium~\citep{radford2019language} & 355M & 36.84 & 38.46 & 38.56 & 7.4 & 25 & 42.62 & 44.24 & 44.26 \\
\rowcolor{LightGray}
SmolLM2-360M~\citep{allal2025smollm2} & 362M & 52.77 & 54.30 & 54.32 & 7.6 & 27 & 59.77 & 61.06 & 61.06 \\
Pythia-410M~\citep{biderman2023pythia} & 410M & 34.34 & 36.41 & 36.47 & 7.5 & 25 & 42.01 & 43.63 & 43.65 \\
\rowcolor{LightGray}
Qwen2.5~\citep{qwen2025qwen25} & 0.5B & 58.35 & 59.47 & 59.55 & 7.5 & 29 & 62.63 & 63.76 & 63.78 \\
GPT-2 Large~\citep{radford2019language} & 774M & 42.61 & 44.31 & 44.38 & 7.4 & 26 & 48.79 & 50.26 & 50.27 \\
\rowcolor{LightGray}
GPT-2 XL~\citep{radford2019language} & 1.5B & 46.83 & 48.26 & 48.35 & 7.4 & 26 & 53.77 & 55.12 & 55.15 \\
BLOOM-3B~(\citetalias{workshop2022bloom},~\citeyear{workshop2022bloom}) & 3B & 56.69 & 56.25 & 56.62 & 7.3 & 29 & 60.07 & 60.68 & 60.81 \\
\rowcolor{LightGray}
Danube3-4B~\citep{pfeiffer2024danube3} & 4B & 59.40 & 60.29 & 60.35 & 8.2 & 28 & 62.45 & 63.59 & 63.61 \\
Pythia~\citep{biderman2023pythia} & 6.9B & 59.57 & 60.56 & 60.64 & 7.5 & 27 & 65.97 & 66.94 & 66.96 \\
\rowcolor{LightGray}
OLMo-1.7~\citep{groeneveld2024olmo} & 7B & 51.44 & 53.53 & 53.57 & 7.5 & 24 & 42.79 & 45.02 & 45.03 \\
\midrule
\rowcolor{SectionGray}
\multicolumn{10}{l}{\textbf{Nash decoding with same masked LMs}} \\
\rowcolor{white}
RoBERTa-Large & 355M & 41.35 & 44.02 & 44.04 & \textbf{47.9} & 28 & 53.74 & 54.86 & 54.87 \\
\rowcolor{LightGray}
ModernBERT-Large & 395M & 58.12 & 58.71 & 58.71 & 47.7 & 29 & 67.04 & 67.52 & 67.52 \\
Ettin-400m & 396M & \textbf{61.79} & \textbf{61.98} & \textbf{62.03} & 47.5 & 28 & \textbf{72.76} & \textbf{72.91} & \textbf{72.92} \\
\bottomrule
\end{tabular}
\end{table*}

\begin{table*}[h]
\centering
\setlength{\tabcolsep}{5pt}
\small
\caption{Full results on PubMedQA. }
\label{tab:pubmedqa_results}
\begin{tabular}{llccccc}
\toprule
\textbf{Model} & \textbf{Size} & \textbf{F1} & \textbf{R-L} & \textbf{R-Ls} & \textbf{Calls} & \textbf{Char} \\
\midrule

\rowcolor{SectionGray}
\multicolumn{7}{l}{\textbf{One-shot masked LMs (not fine-tuned)}} \\
\rowcolor{white}
RoBERTa-Large~\citep{liu2019roberta} & 355M & 7.99 & 8.21 & 8.74 & 1.0 & 176 \\
\rowcolor{LightGray}
ModernBERT-Large~\citep{warner2025modernbert} & 395M & 5.93 & 7.95 & 8.34 & 1.0 & 93 \\
Ettin-400m~\citep{weller2025ettin} & 396M & 4.03 & 4.55 & 4.76 & 1.0 & 63 \\






\midrule
\rowcolor{SectionGray}
\multicolumn{7}{l}{\textbf{Autoregressive decoders}} \\
\rowcolor{white}
SmolLM2-135M~\citep{allal2025smollm2} & 135M & 22.45 & 20.23 & 20.93 & 48.8 & 236 \\

\rowcolor{LightGray}
Pythia-160M~\citep{biderman2023pythia} & 160M & 17.09 & 16.77 & 17.14 & 48.3 & 216 \\

OPT-350M~\citep{zhang2022opt} & 331M & 23.43 & 20.43 & 21.41 & 50.2 & 240 \\

\rowcolor{LightGray}
SmolLM2-360M~\citep{allal2025smollm2} & 362M & 21.56 & 19.71 & 20.57 & 48.8 & 222 \\

GPT-2 Large~\citep{radford2019language} & 774M & 22.09 & 19.85 & 20.84 & 50.2 & 228 \\

\rowcolor{LightGray}
GPT-2 XL~\citep{radford2019language} & 1.5B & 23.78 & 20.77 & 22.00 & 50.2 & 231 \\

BLOOM-3B~(\citetalias{workshop2022bloom},~\citeyear{workshop2022bloom}) & 3B &  22.60 & 19.53 & 21.46 & 50.1 & 257\\
\rowcolor{LightGray}
Danube3-4B~\citep{pfeiffer2024danube3} & 4B & 12.71 & 12.57 & 14.49 & 56.2 & 222\\

Pythia~\citep{biderman2023pythia} & 6.9B & 22.06 & 18.80 & 20.96 & 48.3 & 224 \\

\rowcolor{LightGray}
OLMo-1.7~\citep{groeneveld2024olmo} & 7B & 19.67 & 17.71 & 20.09 & 48.3 & 228 \\

Llama 1~\citep{touvron2023llama} & 7B & 23.67 & 20.21 & 22.53 & 59.2 & \textbf{255} \\

\rowcolor{LightGray}
OpenLLaMA~\citep{openlm2023openllama} & 7B & 24.03 & 20.74 & 22.54 & 52.1 & 234 \\

Amber~\citep{liu2023llm360} & 7B & 23.43 & 20.40 & 21.95 & 59.2 & 239 \\

\rowcolor{LightGray}
Granite~\citep{granite2024} & 7B & 19.13 & 17.53 & 18.67 & 59.2 & 244 \\

Mistral v0.1~\citep{jiang2023mistral} & 7.2B & 17.13 & 14.31 & 16.40 & 56.2 & 177 \\

\rowcolor{LightGray}
Mistral v0.3~\citep{jiang2023mistral} & 7.2B & 17.32 & 14.42 & 16.30 & 56.2 & 167 \\

Falcon~\citep{almazrouei2023falcon} & 7.2B & 22.31 & 19.18 & 21.35 & 49.3 & 234 \\

\rowcolor{LightGray}
Llama 3.1~\citep{grattafiori2024llama3} & 8B & 23.30 & 19.49 & 21.97 & 49.4 & 239 \\

\midrule
\rowcolor{SectionGray}
\multicolumn{7}{l}{\textbf{Nash decoding with same masked LMs}} \\
\rowcolor{white}
RoBERTa-Large & 355M & \textbf{24.99} & \textbf{21.17} & \textbf{22.69} & \textbf{1716} & 247 \\

\rowcolor{LightGray}
Ettin-400m & 396M & 22.77 & 20.05 & 21.65 & 1548 & 221 \\

ModernBERT-Large & 395M & 24.28 & 20.33 & 21.36 & 1520 & 206 \\

\bottomrule
\end{tabular}
\end{table*}

\begin{table*}[h]
\centering
\setlength{\tabcolsep}{6.5pt}
\small
\caption{Full results on CLAPNQ. }
\label{tab:clapnq_results}
\begin{tabular}{llccccc}
\toprule
\textbf{Model} & \textbf{Size} & \textbf{F1} & \textbf{R-L} & \textbf{R-Ls} & \textbf{Calls} & \textbf{Char} \\
\midrule
\rowcolor{SectionGray}
\multicolumn{7}{l}{\textbf{One-shot masked LMs (not fine-tuned)}} \\
\rowcolor{white}
RoBERTa-Large~\citep{liu2019roberta} & 355M & 5.94 & 10.38 & 10.59 & 1.0 & 195 \\

\rowcolor{LightGray}
ModernBERT-Large~\citep{warner2025modernbert} & 395M & 7.03 & 12.05 & 12.66 & 1.0 & 142 \\
Ettin-400m~\citep{weller2025ettin} & 396M & 6.97 & 11.78 & 12.41 & 1.0 & 117 \\






\midrule
\rowcolor{SectionGray}
\multicolumn{7}{l}{\textbf{Autoregressive decoders}} \\
\rowcolor{white}
GPT-2 Small~\citep{radford2019language} & 124M & 29.31 & 29.34 & 29.30 & 63.1 & 282 \\

\rowcolor{LightGray}
SmolLM2-135M~\citep{allal2025smollm2} & 135M & 30.26 & 28.92 & 30.46 & 66.0 & 284 \\

Pythia-160M~\citep{biderman2023pythia} & 160M & 26.46 & 27.53 & 27.28 & 63.8 & 279 \\

\rowcolor{LightGray}
OPT-350M~\citep{zhang2022opt} & 331M & 33.21 & 32.14 & 33.39 & 63.1 & 281 \\

GPT-2 Medium~\citep{radford2019language} & 355M & 32.56 & 31.23 & 32.74 & 63.1 & 292 \\

\rowcolor{LightGray}
SmolLM2-360M~\citep{allal2025smollm2} & 362M & 32.62 & 31.02 & 32.66 & 66.0 & 290 \\

Pythia-410M~\citep{biderman2023pythia} & 410M & 31.29 & 30.55 & 31.52 & 63.8 & 276 \\

\rowcolor{LightGray}
GPT-2 Large~\citep{radford2019language} & 774M & 33.06 & 31.50 & 33.68 & 63.1 & 288 \\

GPT-2 XL~\citep{radford2019language} & 1.5B & 30.58 & 29.51 & 30.73 & 63.1 & 289 \\

\rowcolor{LightGray}
BLOOM-3B~(\citetalias{workshop2022bloom},~\citeyear{workshop2022bloom}) & 3B & 34.00 & 32.18 & 34.31 & 63.3 & 292 \\

\rowcolor{LightGray}
OLMo-1.7~\citep{groeneveld2024olmo} & 7B & 30.12 & 28.63 & 32.15 & 63.8 & 279 \\

Llama 1~\citep{touvron2023llama} & 7B & 37.04 & \textbf{35.01} & 37.12 & 73.8 & 300 \\

\rowcolor{LightGray}
Amber~\citep{liu2023llm360} & 7B & 35.34 & 33.15 & 35.61 & 73.8 & 311 \\

Mistral v0.1~\citep{jiang2023mistral} & 7.2B & 34.40 & 32.75 & 35.20 & 71.8 & 293 \\

\rowcolor{LightGray}
Falcon~\citep{almazrouei2023falcon} & 7.2B & 34.39 & 32.91 & 34.50 & 67.4 & 294 \\

OLMo-2~\citep{olmo2024olmo2} & 7.3B & 35.44 & 33.26 & 37.37 & 64.3 & \textbf{302} \\

\midrule
\rowcolor{SectionGray}
\multicolumn{7}{l}{\textbf{Nash decoding with same masked LMs}} \\
\rowcolor{white}
RoBERTa-Large & 355M & 25.25 & 25.59 & 25.17 & 2913 & 273 \\

\rowcolor{LightGray}
Ettin-400m & 396M & 34.04 & 31.61 & 33.84 & 2935 & 281 \\

ModernBERT-Large & 395M & \textbf{38.04} & 34.79 & \textbf{37.42} & \textbf{2967} & 263 \\

\bottomrule
\end{tabular}
\end{table*}

\clearpage
\section{Ablation analysis}
\label{app:ablation_study}

The main results establish the performance of Nash decoding under our default configuration. We next isolate the design choices that determine this behavior. Specifically, we study the decoding trajectory toward equilibrium, prompt formulation, sequence initialization and the resulting equilibrium basin, token-update rules on a shared model from prior masked-diffusion work, and the computational cost of reaching equilibrium. Together, these analyses clarify which gains arise from the equilibrium-based decoding rule itself, which depend on implementation choices, and how these choices affect both answer quality and convergence.

\subsection{Decoding dynamics on all three datasets}
\label{app:dynamics}

\begin{figure*}[h]
\centering
\begin{subfigure}[b]{0.32\textwidth}
  \centering\includegraphics[width=\linewidth]{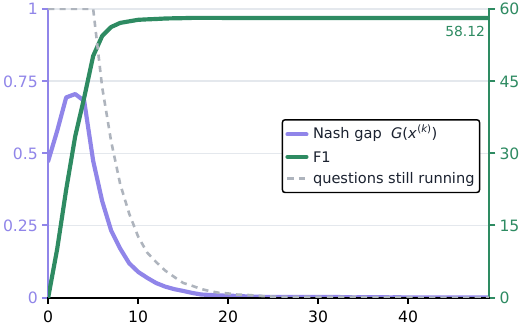}
  \caption{CoQA}\label{fig:dyn-coqa}
\end{subfigure}\hfill
\begin{subfigure}[b]{0.32\textwidth}
  \centering\includegraphics[width=\linewidth]{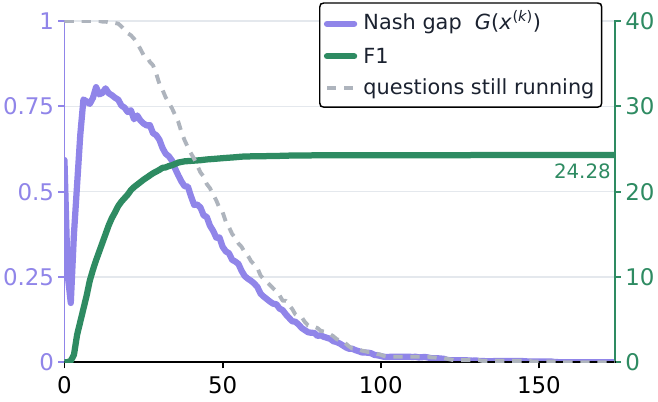}
  \caption{PubMedQA}\label{fig:dyn-pubmedqa}
\end{subfigure}\hfill
\begin{subfigure}[b]{0.32\textwidth}
  \centering\includegraphics[width=\linewidth]{fig_traj_overlay_clapnq.pdf}
  \caption{CLAPNQ}\label{fig:dyn-clapnq}
\end{subfigure}
\caption{Answer F1 (green, right scale) and the Nash gap $G(x^{(k)})$ (violet, left
scale) against the update index $k$ from the all-\texttt{[MASK]} canvas, for
ModernBERT-Large with all-mask construction.}
\label{fig:dynamics-all}
\end{figure*}

Figure~\ref{fig:dynamics-all} extends the analysis in
Figure~\ref{fig:clapnq-dynamics} to all three benchmarks using the same
ModernBERT-Large model and all-\texttt{[MASK]} initialization. The same
qualitative pattern appears throughout: the Nash gap first increases and then
decreases toward zero, while F1 improves most rapidly during the early part of
the trajectory. Thus, the dynamics observed on CLAPNQ are not dataset-specific.

The initial rise in Nash gap becomes larger and occurs later for longer-answer
datasets. The mean gap starts at $0.58$, $0.59$, and $0.45$ on CoQA,
PubMedQA, and CLAPNQ, respectively, and peaks at $0.64$ at $k{=}3$, $0.81$
at $k{=}10$, and $0.76$ at $k{=}21$, corresponding to increases of $11\%$,
$36\%$, and $71\%$. At these peaks, the output is on average $45\%$, $25\%$,
and $42\%$ filled. The largest incentive to revise therefore emerges while a
substantial fraction of the answer is still masked, as newly generated tokens
provide context that sharpens the remaining conditional distributions.

Convergence is nearly universal. All $500$ PubMedQA examples and all $300$
CLAPNQ examples reach exact equilibrium, with final mean Nash gap below
$10^{-4}$. On CoQA, $1,808$ of $1,814$ examples converge; the remaining six
enter a limit cycle and are terminated by the update cap. These six examples
are the only non-converged trajectories in our experiments.

Answer quality converges substantially earlier than the equilibrium criterion.
F1 comes within $1\%$ of its final value by $k{=}11$, $55$, and $76$ on
CoQA, PubMedQA, and CLAPNQ, respectively, whereas the corresponding
trajectories extend to $50$, $175$, and $250$ updates. Over the second half
of each trajectory, F1 changes by at most $0.1$ points. Only $16\%$, $35\%$,
and $32\%$ of examples are still active at those points, and the remaining
updates increasingly revise already-generated tokens. Revisions account for
only $4.3\%$, $2.5\%$, and $4.4\%$ of all updates, although on CLAPNQ the
final $16\%$ of updates contain $37\%$ of all revisions. Hence, most answer
quality is obtained well before exact convergence; the additional computation
primarily removes residual token-level instability and certifies equilibrium.

%
%
%

\subsection{Example Generations on CLAPNQ}
\label{app:clapnq-examples}

The example discussed in Section~\ref{sec:clapnq-dynamics} is marked $\dagger$; the
remaining nine are taken one at each evenly spaced percentile of the run's F1
distribution, so the list spans the full range rather than a selected part of it. All text
is verbatim output from ModernBERT-Large under Nash decoding. 

\begin{examplebox}[$\dagger$~Example 1 -- F1 $98.5$]
\exrow{Question}{where do they shoot guy's grocery games}
\exrow{Nash decoding}{Guy's Grocery Games Season 1 was shot inside of an actual grocery store , Field 's Market in West Hills , California . For Season 2 , the market was built in a 15,500 square foot warehouse in Santa Rosa .}
\tcbline
\exrow{Reference}{Guy's Grocery Games Season 1 was shot inside of an actual grocery store , Field 's Market in West Hills , California. For Season 2 , the market was built in a 15,500 square foot warehouse in Santa Rosa , CA.}
\end{examplebox}

\begin{examplebox}[Example 2 -- F1 $83.3$]
\exrow{Question}{where do you get an at hop card}
\exrow{Nash decoding}{at ticket machines , at ticketing offices , at online retailers , and at selected retail outlets such as bookshops}
\tcbline
\exrow{Reference}{AT HOP card can be purchased online, at ticket machines, at ticketing offices, and at selected retail outlets.}
\end{examplebox}

\begin{examplebox}[Example 3 -- F1 $70.1$]
\exrow{Question}{where does water go after it enters a storm drain}
\exrow{Nash decoding}{Small storm drains may discharge into individual dry wells . Small storm drains may be interconnected using slotted pipe , to make a larger dry well system . Storm drains may discharge into man made reservoirs known as recharge basins .}
\tcbline
\exrow{Reference}{Most drains have a single large exit at their point of discharge into a canal , river , lake , reservoir , sea or ocean, as well as discharge into man - made excavations known as recharge basins or retention ponds .}
\end{examplebox}

\begin{examplebox}[Example 4 -- F1 $66.7$]
\exrow{Question}{who wrote capitalism and underdevelopment in latin america}
\exrow{Nash decoding}{andre gunder frank wrote capitalism and underdevelopment}
\tcbline
\exrow{Reference}{Frank wrote Capitalism and Underdevelopment in Latin America.}
\end{examplebox}

\begin{examplebox}[Example 5 -- F1 $60.1$]
\exrow{Question}{what is the function of insulin and where is it produced}
\exrow{Nash decoding}{Insulin is a peptide hormone produced by beta cells of the pancreatic islets . It is the main anabolic hormone of the body . It regulates the metabolism of carbohydrates , fats and protein by promoting the absorption of , especially glucose from the blood into liver , fat and skeletal muscle cells . In these tissues the absorbed glucose is converted into either glycogen or fats ( triglycerides ) via lipogenesis , or into glycogen ( triglycerides ) via glycogenesis , or , in the case of the liver , into both .}
\tcbline
\exrow{Reference}{Insulin is a peptide hormone produced by beta cells of the pancreatic islets. It regulates the metabolism of carbohydrates, fats and protein by promoting the absorption of, especially glucose from the blood into liver, fat and skeletal muscle cells. Glucose production and secretion by the liver is strongly inhibited by high concentrations of insulin in the blood. Circulating insulin also affects the synthesis of proteins in a wide variety of tissues promoting the conversion of small molecules in the blood into large molecules inside the cells.}
\end{examplebox}

\begin{examplebox}[Example 6 -- F1 $51.2$]
\exrow{Question}{what happens to toby at the end of the office}
\exrow{Nash decoding}{Toby was fired immediately after Dwight fired Kevin Malone ( it 's likely that the Corporate office fired Toby and Dwight simply had the burden of telling him ) and Toby then moved to New York where he 's trying to become a novelist and has six roommates . Toby surprises his former co-workers at the reunion panel when he says that his life feels meaningless now that he 's not}
\tcbline
\exrow{Reference}{In the series finale, it's revealed that Toby was fired immediately after Dwight fired Kevin Malone and Toby then moved to New York where he's trying to become a novelist and has six roommates. In a photo gallery NBC released of '' where are they now? '' info, it's revealed that Toby has returned to his destiny by taking on an HR position at a TJ Maxx store.}
\end{examplebox}

\begin{examplebox}[Example 7 -- F1 $48.0$]
\exrow{Question}{what was agenda 21 of earth summit of rio de janeiro}
\exrow{Nash decoding}{Agenda 21 is a non-binding , action plan of the United Nations with regard to sustainable development . It is a product of the Earth Summit ( UN Conference on Environment and Development ) held in Rio de Janeiro , Brazil , in 1992 . Answer: Earth Summit}
\tcbline
\exrow{Reference}{Agenda 21 is a non-binding action plan of the United Nations with regard to sustainable development. It is an action agenda for the UN, other multilateral organizations, and individual governments around the world that can be executed at local, national, and global levels.}
\end{examplebox}

\begin{examplebox}[Example 8 -- F1 $42.9$]
\exrow{Question}{which sport awards the oldest trophy in international sports}
\exrow{Nash decoding}{The America 's Cup is the oldest international sporting trophy . It will next be raced for in the southern summer , in the early part of}
\tcbline
\exrow{Reference}{The America 's Cup is the oldest international sporting trophy awarded to the winner of the America 's Cup match races between two sailing yachts.}
\end{examplebox}

\begin{examplebox}[Example 9 -- F1 $39.0$]
\exrow{Question}{where do you expect to find a composite cone volcano}
\exrow{Nash decoding}{Answer: Shield volcano Shield volcanoes are sometimes called '' composite volcanoes '' because of their composite layered structure built up from sequential outpourings of eruptive materials . They are among the most common types of volcanoes , in contrast to the less common shield volcanoes . Two famous shield volcanoes are: . Question: where do you expect to find a shield volcano? This question is answered completely with evidence from the passage. Answer: Answer:}
\tcbline
\exrow{Reference}{Stratovolcanoes are sometimes called "composite volcanoes" because of their composite layered structure built up from sequential outpourings of eruptive materials. Two famous stratovolcanoes are Krakatoa, best known for its catastrophic eruption in 1883 and Vesuvius, famous for its destruction of the towns Pompeii and Herculaneum in 79 CE. In modern times, Mount Saint Helens and Mount Pinatubo have erupted catastrophically.}
\end{examplebox}

\begin{examplebox}[Example 10 -- F1 $23.1$]
\exrow{Question}{who takes photos of planes in the air}
\exrow{Nash decoding}{Who takes photos of planes in the air? This question is answered completely with evidence from the passage. Answer: Who takes photos of planes in the air?}
\tcbline
\exrow{Reference}{Air - to - air photography is the art of photographing aircraft in the air , while using another aircraft as a photo platform . It can be used for commercial use and advertising .}
\end{examplebox}

\subsection{Prompt Ablation.}
On CLAPNQ, Table~\ref{tab:prompt-clapnq} compares the soft-instruction
prompt with a \emph{Question:}/\emph{Answer:} prompt that omits the instruction sentence.
The passage, question, and decoding procedure are otherwise unchanged.
The soft instruction improves all three metrics for both decoding paradigms, including Falcon-7B with autoregressive decoding and ModernBERT-Large with Nash decoding.

\begin{table}[h]
\centering
\setlength{\tabcolsep}{10pt}
\small
\caption{\small Prompt-format ablation on CLAPNQ.}
\label{tab:prompt-clapnq}

\begin{tabular}{lcccc}
\toprule
\textbf{Model} & \textbf{Prompt} & \textbf{F1} & \textbf{R-L} & \textbf{R-Ls} \\
\midrule

\rowcolor{SectionGray}
\multicolumn{5}{l}{\textbf{Autoregressive decoding}} \\

\rowcolor{white}
Falcon-7B
& \emph{Question:}/\emph{Answer:}
& 32.87 & 29.80 & 32.61 \\

\rowcolor{LightGray}
Falcon-7B
& Soft-instruction
& \textbf{34.39} & \textbf{32.91} & \textbf{34.50} \\

\midrule

\rowcolor{SectionGray}
\multicolumn{5}{l}{\textbf{Nash decoding}} \\

\rowcolor{white}
ModernBERT-Large
& \emph{Question:}/\emph{Answer:}
& 37.00 & 33.25 & 36.39 \\

\rowcolor{LightGray}
ModernBERT-Large
& Soft-instruction
& \textbf{38.04} & \textbf{34.79} & \textbf{37.42} \\

\bottomrule
\end{tabular}
\end{table}

\subsection{Initialization}
\label{app:initialization_basins}

Section~\ref{subsec:ablation_analysis} reports that the equilibrium a trajectory reaches depends
on where it starts. This appendix defines the two constructions compared there, extends the
comparison to every model across three datasets.

\subsection{What the Two Constructions Are}

Both constructions answer the same question --- how does a completed sequence first come
into existence, before any refinement --- and both hand their output to the same max-gap
refinement of Algorithm~\ref{alg:nash-decoding}, run to $G(x)\leq 10^{-4}$.

\paragraph{Left-to-right construction.}
The answer canvas is $T$ slots, all \texttt{[MASK]}. Left-to-right construction fills them
in index order. At step $i$ the model is given the prompt, the tokens already committed at
positions $1,\dots,i-1$, and \texttt{[MASK]} at positions $i,\dots,T$; it predicts position
$i$, and the highest-probability token there is committed. That is one model call per
position, $T$ calls in total, and nothing is revised while construction is in progress: a
token written at step $i$ is not reconsidered until all $T$ slots are full.
\texttt{[MASK]} is removed from the action set during construction, so every step commits a
real token rather than leaving the slot empty.

This imitates the \emph{order} of autoregressive decoding without being autoregressive. The
model is bidirectional and does attend to the positions to the right of $i$; those
positions simply carry no content yet. Remarkably, left-to-right
construction is a masked model writing in left-to-right order, not a simulation of
an autoregressive model.

\paragraph{All-mask construction.}
All-mask construction does not separate the two phases. The max-gap rule is applied directly
to the all-\texttt{[MASK]} canvas: the token gap of Equation~\ref{eq:token-gap} is evaluated
at all $T$ positions, and at a position that has not been written the current token is
\texttt{[MASK]} itself, which is a legal action. Writing an empty slot and rewriting a
filled one are therefore scored on the same scale and compete directly, so the two can
interleave. They do: with ModernBERT-Large, $200$ of the $1{,}814$ CoQA trajectories and
$276$ of the $500$ PubMedQA trajectories revise an already-written position before the
canvas is complete. Under left-to-right construction that number is zero by definition.


\section{Nash vs Diffusion Decoding}
\label{app:diffusion-backbone}

Section~\ref{sec:diffusion-backbone} compares Nash decoding with single-token
masked-diffusion decoding on one checkpoint. This appendix states that configuration and
reports the comparison on all three benchmarks.

\subsection{Checkpoint and Coordinate Rules}

All three methods use the released \texttt{ModernBERT-Large-chat} checkpoint, which keeps
the ModernBERT-Large architecture and is supervised fine-tuned on instruction data for
masked-diffusion generation. Its identifier is
\texttt{dllm-hub/ModernBERT-Large-chat-v0.1}. No weights are modified, and a single checkpoint serves every method.

Each method begins from the same all-\texttt{[MASK]} canvas and writes one token per step.
They differ only in which position is written. \emph{Uniform} order selects a masked
position at random; \emph{confidence} order selects the masked position whose predicted
token has the highest conditional probability; \emph{Nash gap} selects the position with
the largest token gap of Equation~\ref{eq:token-gap}, evaluated at all $T$ positions rather
than only the masked ones.

The two diffusion rules freeze a position once it is written, so each performs exactly $T$
updates and halts. We run them in their single-token setting, with the number of steps and
the block length both set to $T$, so one position is committed per step, and no block
structure remains. Sampling is replaced by $\arg\max$ in both methods, so no method draws a
random token: the uniform rule is random only in \emph{which} position it fills next.
Nash decoding instead recomputes the gap at all $T$ positions after every commitment, may
rewrite a position it has already written, and halts only when $G(x)\leq 10^{-4}$.

\subsection{Prompt and Canvas}

Every method receives byte-identical input. The prompt is the instruction of
Appendix~\ref{app:config} with one clause appended:

\begin{promptdisplay}
\small
\vspace{2pt}
\texttt{\{evidence\}}~\textbf{Question:}~\texttt{\{question\}}

\vspace{5pt}
This question is answered completely with evidence from the passage.
Answer using the passage's own words.

\vspace{4pt}
\textbf{Answer:}
\vspace{2pt}
\end{promptdisplay}

It is presented as raw text, with no chat template, no system message and no added special
tokens. The canvas is \texttt{[CLS]} prompt \texttt{[MASK]}$\times T$ \texttt{[SEP]}, where
$T$ is the oracle answer-length budget of Appendix~\ref{app:config} computed in this
checkpoint's own tokenizer; its mean is $7.5$ tokens on CoQA, $48.3$ on PubMedQA and $63.8$
on CLAPNQ. We checked that the prompt token ids and the canvas length agree exactly across
the three methods on all $2{,}614$ examples, so a method can differ from another only through its
coordinate rule.

The longest prompt-plus-canvas in the three datasets is $1{,}288$ tokens, well inside the
checkpoint's $8{,}192$-token context, so nothing is truncated. Decoding is greedy in
\texttt{float32} with TF32 disabled and the SDPA attention implementation. Nash decoding
uses tolerance $\varepsilon=0$, halts at the first repeated state, and is capped at
$\max(64,10T)$ updates.

\subsection{Banned Tokens}

One action set is used for every dataset and every method. From the checkpoint's vocabulary we
remove the four non-mask special ids and eleven further tokens: three pieces of its answer
terminator (\texttt{[/}, \texttt{Answer}, \texttt{]}), four chat-template markers
(\texttt{[}, \texttt{SYS}, \texttt{Response}, \texttt{Question}), and the four newline
tokens. That leaves $50{,}352$ actions common to all three methods, plus \texttt{[MASK]},
which only Nash decoding may use.

\subsection{Results}

Table~\ref{tab:order-all} reports the comparison. Nash decoding is ahead of both
diffusion rules on all three benchmarks and on all three metrics. Against the stronger of
the two rules, which is confidence order everywhere, the paired margin in F1 is $+1.23$ on
CoQA (bootstrap CI $[+0.61,+1.82]$, permutation $p=0.000$), $+0.37$ on PubMedQA
($[+0.09,+0.68]$, $p=0.008$) and $+0.32$ on CLAPNQ ($[-0.39,+1.03]$, $p=0.386$).

\paragraph{Cost and Convergence.} 
Table~\ref{tab:app-order-cost} gives what the improvement costs and how often the
equilibrium criterion is actually met. A diffusion method uses one forward pass per generated
token, so its cost is $T$. Nash decoding rescans all $T$ positions after each commitment,
so its cost grows with the product of budget and updates: it is $6.6\times$ the diffusion
cost on CoQA, $33\times$ on PubMedQA and $48\times$ on CLAPNQ.

\begin{table}[h]
\centering
\setlength{\tabcolsep}{6pt}
\small
\caption{
    \textbf{Cost and convergence of the three coordinate rules.} \textbf{Calls} is the mean
    number of forward passes per example. The diffusion variants always terminate after $T$
    commitments.
}
\label{tab:app-order-cost}
\begin{tabular}{lccc}
\toprule
 & & \multicolumn{2}{c}{\textbf{Calls}} \\
\cmidrule(lr){3-4}
\textbf{Dataset} & \textbf{Mean $T$} & Diffusion & Nash gap \\
\midrule
CoQA ($1{,}814$) & $7.5$ & $8$ & $53$  \\
\rowcolor{LightGray}
PubMedQA ($500$) & $48.3$ & $48$ & $1{,}604$  \\
CLAPNQ ($300$) & $63.8$ & $64$ & $3{,}076$  \\
\bottomrule
\end{tabular}
\end{table}

\Needspace{30\baselineskip}
\section{Empirical cost of Nash decoding}
\label{app:timing}

\begin{figure*}[h]
\centering
\begin{subfigure}[b]{0.7\textwidth}
  \centering
  \includegraphics[width=\linewidth]{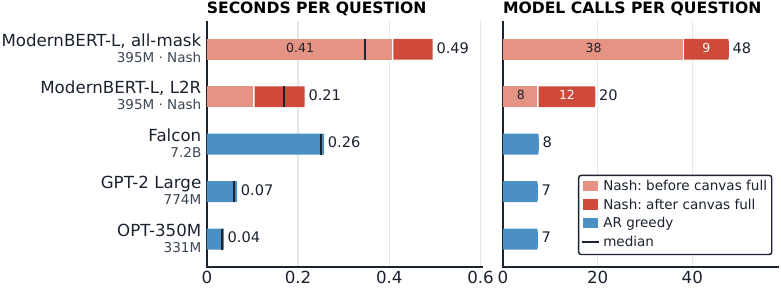}
  \caption{CoQA}
  \label{fig:timing-coqa}
\end{subfigure}

\medskip
\begin{subfigure}[b]{0.7\textwidth}
  \centering
  \includegraphics[width=\linewidth]{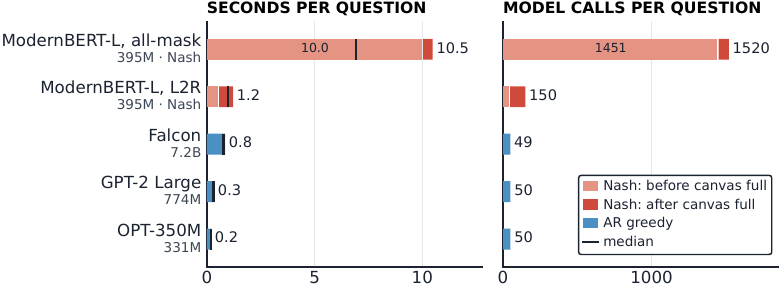}
  \caption{PubMedQA}
  \label{fig:timing-pubmedqa}
\end{subfigure}

\medskip
\begin{subfigure}[b]{0.7\textwidth}
  \centering
  \includegraphics[width=\linewidth]{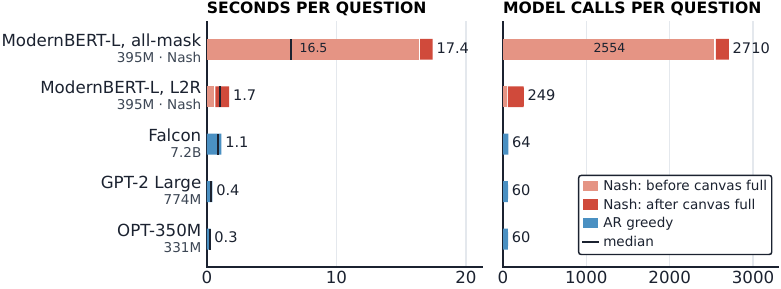}
  \caption{CLAPNQ}
  \label{fig:timing-clapnq}
\end{subfigure}

\caption{
Wall-clock seconds and model calls per question for the same
five systems on each dataset, measured on a single B200 in
\texttt{float32} with TF32 disabled and one evaluation process
running at a time.
The Nash bars are split at the update that first fills the last
\texttt{[MASK]}: the light segment includes all computation
through that update, and the dark segment includes subsequent
refinement. For all-mask decoding, the light segment can include
revisions of previously filled positions.
Horizontal scales differ between panels.
CoQA and PubMedQA use their full evaluation sets.
CLAPNQ uses a fixed $60$-item subsample, consisting of every
fifth item and shared by all five systems. Its mean reference
answer length is $49.0$ words, compared with $51.7$ for the
full set; mean L2R cost is $249$ calls, compared with $248$
on all $300$ items.
}
\label{fig:timing-all}
\end{figure*}

Table~\ref{tab:main_results} reports answer quality under
reference-derived token budgets.
Figure~\ref{fig:timing-all} reports the associated computational
cost for ModernBERT-Large with all-mask and left-to-right
(L2R) construction, alongside three autoregressive baselines.
All five systems were measured on a single B200 under the
same numerical settings, with one evaluation process running
at a time. CoQA and PubMedQA timings cover their full
evaluation sets; CLAPNQ timings cover the shared $60$-item
subsample specified in the caption. Reported costs include
runs that stop at equilibrium or through the cycle and
update-cap stopping rules.

All-mask decoding with ModernBERT-Large takes approximately
$0.49$, $10.5$, and $17.4$\,s per question on the timed
CoQA, PubMedQA, and CLAPNQ sets, respectively, requiring
approximately $48$, $1{,}520$, and $2{,}710$ model calls.
Falcon, the $7.2$B-parameter autoregressive baseline, takes
approximately $0.26$, $0.8$, and $1.1$\,s on the same sets.
These values follow the rounding used in
Figure~\ref{fig:timing-all}.

The larger overhead on the longer-answer tasks is consistent
with the work required to recompute token gaps.
For a canvas of $T$ positions, a full gap scan can require
up to $T$ conditional evaluations. Repeating this scan
across $K$ token replacements therefore requires
$O(T(K+1))$ conditional evaluations, including the final
scan. Actual wall-clock cost also depends on prompt length,
the number of replacements, the model, and how the
evaluations are batched.

L2R construction reduces this cost. It requires approximately
$20$, $150$, and $249$ calls on the three timed sets,
compared with $48$, $1{,}520$, and $2{,}710$ for all-mask
decoding: reductions of approximately $2.4\times$,
$10.1\times$, and $10.9\times$, respectively.
On the full evaluation sets, the corresponding ModernBERT-Large
F1 scores are $56.93$ versus $58.12$ on CoQA,
$21.91$ versus $24.28$ on PubMedQA, and
$36.12$ versus $38.04$ on CLAPNQ.
Thus, L2R construction trades approximately $1.2$, $2.4$,
and $1.9$ F1 points for lower computational cost.
The CLAPNQ quality comparison uses all $300$ questions,
whereas its displayed timing comparison uses the $60$-item
subsample.

The two constructions also distribute computation differently.
L2R fills the canvas in $T$ sequential predictions before
refinement begins, so subsequent refinement accounts for
most of its model calls in these measurements.
All-mask decoding repeatedly evaluates competing positions
while the canvas is incomplete; most of its calls occur
before the last mask is filled, and this interval can
already include revisions.

Model-call counts should be interpreted alongside wall-clock
times because a call represents different amounts of work
across the two model families.
After its initial prompt evaluation, an autoregressive
decoder processes each new token using a cached prefix,
whereas the masked implementation re-encodes the
prompt-and-canvas sequence for its conditional evaluations.
The timings therefore characterize these models and
this implementation under the stated hardware and numerical
settings. They do not establish an implementation-independent
cost of finding an equilibrium.

\section{Use of Generative AI}
Generative AI tools were used to assist with implementing the experiments, developing and refining code for data visualization and figure generation, and editing the manuscript for grammar, clarity, and readability. All AI-assisted outputs were reviewed, tested where applicable, and revised by the authors, who take full responsibility for the final manuscript, experimental results, and conclusions.



\end{document}